\documentclass{article}

\usepackage[nonatbib, final]{ap_2023}

\usepackage[utf8]{inputenc} 
\usepackage[T1]{fontenc}    
\usepackage{hyperref}       
\usepackage{url}            
\usepackage{booktabs}       
\usepackage{amsfonts}       
\usepackage{nicefrac}       
\usepackage{microtype}      
\usepackage{biblatex}
\usepackage{multirow}
\usepackage{graphicx}
\usepackage{xcolor}
\usepackage{adjustbox}
\usepackage{tabularx}
\usepackage{caption}
\usepackage{subcaption}
\usepackage{multirow}
\usepackage{svg}

\usepackage{caption,subcaption}

\title{Alquist 5.0: Dialogue Trees Meet Generative Models. A Novel Approach for Enhancing SocialBot Conversations}

\author{
  Ond\v{r}ej Kobza, Jan \v{C}uhel, Tommaso Gargiani, David Herel, Petr Marek, \\Jan Pichl, Jakub Konrad, Petr Lorenc\\
  Faculty of Electrical Engineering, CTU in Prague\\
  Prague, Czechia \\
  \texttt{\{kobzaond, cuheljan, gargitom, hereldav, marekp17\}@fel.cvut.cz} \\
  \texttt{\{pichljan, konrajak, lorenpe2\}@fel.cvut.cz} \\
  \And
   Jan \v{S}ediv\'{y} \\
   CIIRC, CTU in Prague \\
   Prague, Czechia \\
   \texttt{jan.sedivy@cvut.cz} \\
}

\begin{document}

\maketitle

\begin{abstract}
We present our SocialBot -- Alquist~5.0 -- developed for the Alexa Prize SocialBot Grand Challenge~5. Building upon previous versions of our system, we introduce the NRG Barista and outline several innovative approaches for integrating Barista into our SocialBot, improving the overall conversational experience. Additionally, we extend our SocialBot to support multimodal devices. This paper offers insights into the development of Alquist~5.0, which meets evolving user expectations while maintaining empathetic and knowledgeable conversational abilities across diverse topics.
\end{abstract}

\section{Introduction}
This paper introduces our SocialBot -- Alquist~5.0 -- created for the Alexa Prize SocialBot Grand Challenge 5 \cite{Johnston2023}.
This university competition aims to create an Alexa skill capable of conversing coherently and engagingly with humans on popular topics and news events for 20 minutes and achieve an average rating of at least 4.0/5.0. This is an extraordinarily challenging task, as we first had to identify the key ingredients that make up a high-quality conversation and then integrate them into our SocialBot. Therefore, we have focused on creating a SocialBot that is both empathetic and knowledgeable about a large number of topics.

Alquist 5.0 is built upon its previous versions created for the past editions of the SocialBot Grand Challenge. 
However, since the last challenge, state-of-the-art conversational AI has made tremendous progress. This advancement has led to increased user expectations.
The release of OpenAI's ChatGPT\footnote{\url{https://openai.com/blog/chatgpt}} in November 2022 has sparked great interest in large language models (LLMs) among the general public by showcasing their capabilities.
Thankfully, in contrast to closed-source models like ChatGPT and Google's Bard\footnote{\url{https://bard.google.com/}}, we have also witnessed the release of many open-source LLMs that reach the performance of their commercial counterparts. This period of prosperity for language modeling has allowed us to shift our focus toward creating a new Neural Response Generator (NRG) -- Barista (see Section\ref{section:generative_models}) -- for our SocialBot.

When integrating Barista into Alquist, we aimed to find the balance between the scripted dialogues present from our SocialBot's past versions and Barista's generation capabilities. Hence, we have developed several innovative approaches for incorporating Barista into our SocialBot, improving the overall conversational experience. We expand on the topic in Section~\ref{section:DM_LLMs}.

Furthermore, in this year's SocialBot Grand Challenge, we built our SocialBot for multimodal devices, i.e. Alexas equipped with a touchscreen. Since a great conversational experience is closely linked with an excellent user interface, we utilize APL\footnote{\url{https://developer.amazon.com/en-US/docs/alexa/alexa-presentation-language/add-visuals-and-audio-to-your-skill.html}} templates combined with our implementations of additional features. See Section\ref{section:UI} for more details.

\section{System Architecture}
\begin{figure}
    \centering
    \includegraphics[width=0.85\textwidth]{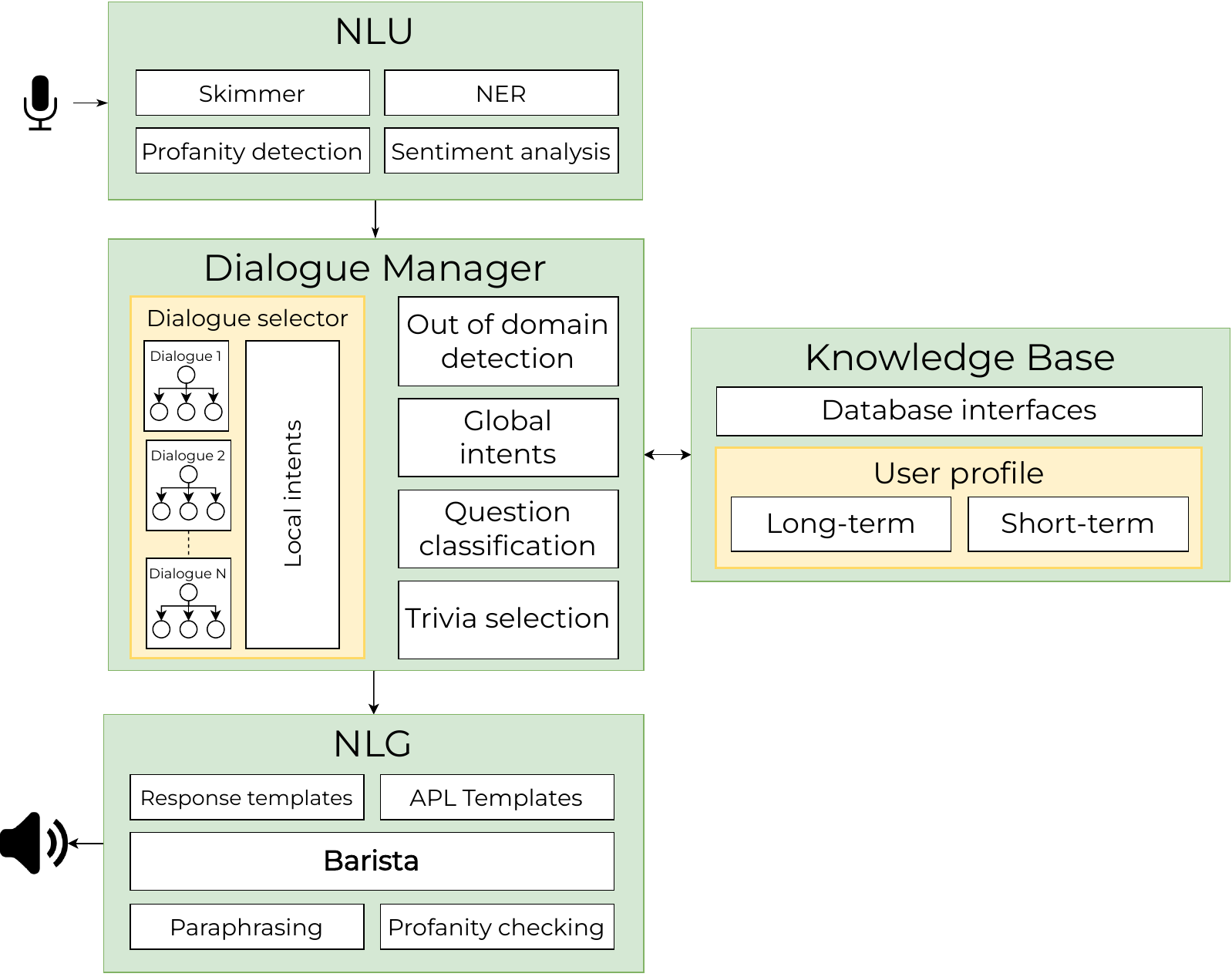}
    \caption{The system architecture builds on top of the work proposed by Konrád~et~al.~\cite{konrad2021alquist}, with the main emphasis being put on the Barista neural response generator.}
    \label{fig:system_architecture}
\end{figure}

The system is based on the architecture proposed by Konrád~et~al.~\cite{konrad2021alquist} (see \autoref{fig:system_architecture}). The main improvement we propose to the architecture is its extension with the NRG Barista (see Section~\ref{section:generative_models}). Barista is a neural response generator designed for open-domain conversations that accesses external knowledge sources to provide knowledgeable responses. We utilize Barista in various conversational contexts described in Section~\ref{section:DM_LLMs}.

From the dialogue design perspective, the main building block of proposed SocialBot is the \emph{main dialogue}. Within the main dialogue, we have a collection of dialogue groups composed of multiple dialogues. Each group is dedicated to a specific topic, such as sports, movies, or music. This design allows our SocialBot to engage in conversations about a wide range of subjects, providing a comprehensive and diverse user experience.
Furthermore, each dialogue follows a tree graph structure. This means that we have the flexibility to specify a conversational flow, defining how the system should respond to different user utterances. By organizing the dialogue in this structured manner, we can ensure a coherent and intuitive conversation experience for the users.

In order to effectively understand user utterances, we follow the practice established in Konrád~et~al.~\cite{konrad2021alquist} and classify every utterance into an \emph{in-domain} or \emph{out-of-domain} intent.
In-domain intents are designed to handle user utterances that are anticipated within the context of the dialogue. Furthermore, we distinguish between two types of in-domain intents: \emph{local}, which are valid only in a specific context and cover the expected user inputs that align with the predefined conversational flow, and \emph{global}, which are valid across all dialogues and can be relevant in any part of the conversation.
On the other hand, out-of-domain intents handle unexpected user utterances that fall outside the predefined scope of the dialogue. Out-of-domain intents play a crucial role when incorporating our new NRG Barista into the system.

We also employ Skimmer, a component introduced by Konrád~et~al.~\cite{konrad2021alquist}. It analyzes user utterances and identifies relevant information. For example, if a user mentions that they have a dog, Skimmer can detect and extract this information without asking the user about it directly.
This ability to gather user details implicitly helps in building the User Profile, another feature introduced by Konrád~et~al.~\cite{konrad2021alquist}. The User Profile is designed to store information about the user, useful for conversation (e.g. that they own a dog), which can be leveraged to provide personalized responses in subsequent interactions.

As in Konrád~et~al.~\cite{konrad2021alquist}, we navigate the user between the Dialogue Trees utilizing a component called Dialogue Selector, which chooses the most relevant topic for conversation. This selection is based on two factors. The first factor is the User Profile, which includes details gathered through Skimmer about the user's preferences, hobbies, and other relevant information. This enables our SocialBot to suggest topics that align with the user's interests. The second factor is the topics identified during the previous interaction with the bot. These topics, detected using entity recognition, are then selected as possible candidates for conversation.

Furthermore, we enrich the dialogues by inserting fun-facts (i.e. interesting trivia information) between dialogues, as described in Konrád~et~al.~\cite{konrad2021alquist}. Some dialogues also use external knowledge bases, specifically Amazon Evi for question answering and The Movie Database\footnote{\url{https://developer.themoviedb.org/docs}} (TMDB) to provide information about movies mentioned during the conversation.

\section{Barista Neural Response Generator}
\label{section:generative_models}
We introduce our innovative Neural Response Generator (NRG) to improve the handling of out-of-domain conversations and its versatility in different dialogue contexts. Our approach begins with the BlenderBot 3 (BB3) model. We developed a novel NRG called Barista, inspired by its architecture.

BlenderBot 3 \cite{bb3}, a 3-billion-parameter model based on the BART \cite{bart} architecture, excels in numerous tasks, such as generating coherent conversations and querying external knowledge sources. Nevertheless, its complexity imposes significant demands on computational resources and efficiency and exhibits various performance-related drawbacks. To address these issues, we propose Barista, an NRG based on BlenderBot 3. Barista is designed to execute tasks in parallel, reducing the need for extensive computational power while enhancing performance. Moreover, it effectively remedies BB3's shortcomings, such as repetitive responses, and improves the overall conversational experience. In this paper, we present an in-depth analysis of Barista and highlight its key improvements to the BB3 architecture.

\subsection{Original BlenderBot 3 Architecture}
The BB3 system utilizes a solitary generative model to execute all tasks inherent to its architecture. Initially, the system assesses three key decisions: whether to incorporate external knowledge, whether to use its long-term memory -- where it archives summaries of user responses in a database as vector embeddings across multiple sessions, and whether to derive a dominant entity, which refers to the primary subject of the current dialogue. 

When BB3 opts for the utilization of its long-term memory, it retrieves a summary from the memory database via DPR~\cite{dpr}. In instances where external knowledge is necessitated, BB3 formulates a search query based on the user's input, input that is then routed to their chosen search engine, Mojeek~\footnote{\url{https://www.mojeek.com/}}. Mojeek, in turn, produces $k$ relevant documents, from which BB3 applies FiD \cite{fid} to distill specific knowledge into a concise form.

Integration of the long-term memory, acquired knowledge, and dominant entity with the user's input and the conversational context is carried out within the BB3 generative model, culminating in the generation of the final dialogue. Subsequently, a summary is formulated based on the user's input and BB3's response, which is then stored for future reference in its long-term memory.

\subsection{Barista Architecture}
\begin{figure}
    \centering
    \includegraphics[width=12cm]{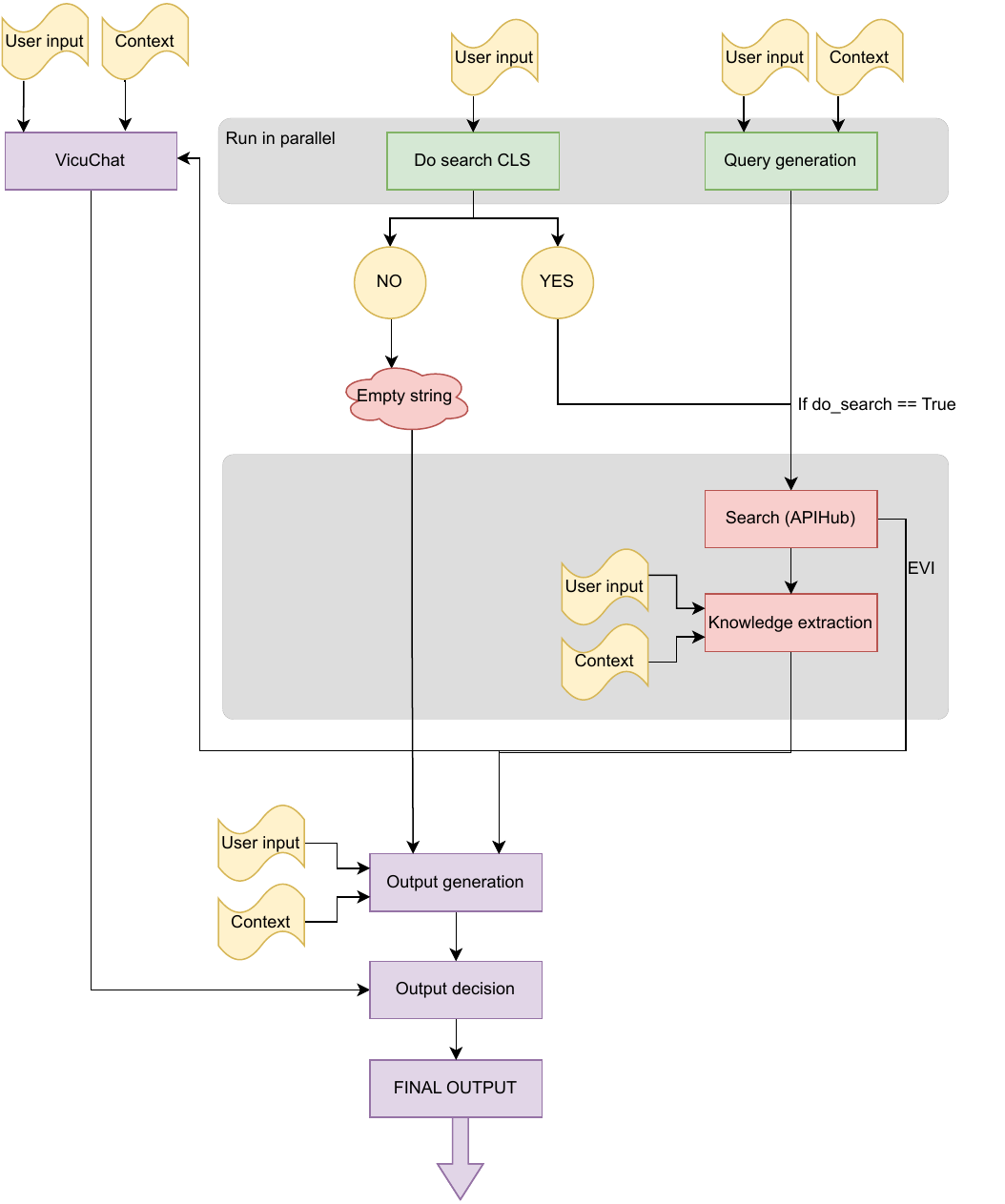}
    \caption{Visualization of Barista.}
    \label{fig:diagram}
\end{figure}

Figure~\ref{fig:diagram} illustrates the architecture of Barista. The pipeline begins with a classification task determining whether the NRG should query external knowledge (long-term memory retrieval and entity extraction are omitted -- \ref{memory}, \ref{ee}).
Concurrently, the system generates a search query for our search API (see Section~\ref{apihub}). Suppose the output of the classifier is $1$ (`yes'). In that case, Barista retrieves information through our search API. In terms of search, the pipeline extracts specific information from a set of documents returned by the API. Subsequently, all external information, user input, and conversational context are fed into the Barista 3B BB3-based model, and the final utterance is generated.


In parallel, Barista runs VicuChat NRG, our fine-tuned version of Vicuna 7B \footnote{\url{https://github.com/lm-sys/FastChat}}. VicuChat generates a response according to the user input, context, and possibly knowledge (together with the query) and generates the next utterance. Finally, a rule-based classifier decides which generated output to use.

\subsection{Novel Models}
To enhance speed and performance, we replaced the original BlenderBot 3 model with our custom models on several subtasks as well as on the main generation task and altered the original architecture. 

\subsubsection{Training and Evaluation Datasets}
Regarding training and evaluation datasets, we employed the same data that was utilized for BB3, specifically for these tasks: Do Search, Do Access User Profile (long-term memory), Query Generation and Knowledge Extraction.

Shuster~et~al.~\cite{bb3} provide a detailed description of the methodology employed in the creation of each BB3 dataset, which encompassed various data mergers and post-processing techniques. The data can be accessed via the ParlAI library\footnote{\url{https://parl.ai/}} that furnishes a platform for downloading and modifying any downloaded data. It is imperative to note that BB3 exploited a distinct set of datasets along with specific modifications for every subtask.
Upon analysis, we identified that certain original training and evaluation datasets displayed imbalance. This discrepancy necessitated the creation of a new test set \ref{testD} for each specific task in order to uncover potential accuracy issues and more effectively assess the performance of our models and the BB3 model. The outcome derived from our test data, and that of the BB3 test data (further called `eval datasets'), indicate that we successfully mitigated the imbalances in the training data through a weighting methodology applied during the training process.\footnote{To differentiate between the BB3 model and the overall system, we refer to the system as the `pipeline' and the model as the `BB3 model' or `3B model'.}

\subsubsection{Test Datasets} \label{testD}
The generation of each test dataset was accomplished in three sequential steps:
\begin{enumerate}
    \item {\textbf{Creation of the Initial Seed:}} During testing particular BB3 modules, we observed manually the inputs and corresponding outputs for each submodule. Notably, these inputs originated from the team members and other students of the Czech Technical University. During these observations, several problematic patterns were identified, and related user utterances were subsequently used to formulate an initial seed.
    \item {\textbf{Expansion of the Initial Seed:}} The initial seed was manually amplified to include a larger number of examples. The objective here was not only to address the few recognized problematic patterns, but also to encapsulate a more diverse set of cases.
    \item {\textbf{Expansion via ChatGPT:}} The third and final step involved leveraging the advanced capabilities of OpenAI's ChatGPT \footnote{\url{https://openai.com/}} to further expand the dataset refined in Stage 2. We provided the model with ten manually selected representative samples from previous stages and asked ChatGPT to generate $n$ more samples \ref{dataPrompts}. The total size of each test set is 300.
\end{enumerate}

\subsubsection{Do Search Classification}
This module informs the pipeline whether to query external knowledge. 
Our best models achieve an improvement of up to 22 percentage points over the original BB3 model and are over 15x (up to 192x) faster. Comprehensive results can be found in Section~\ref{results}.

\subsubsection{Do Access User Profile Classification} \label{memory}
For this task, BB3 used partly synthetic training and evaluation datasets. However, both are heavily unbalanced, with over 90\% negative samples. 
We attempted to optimize smaller models like BERT-tiny~\cite{tinybert} and ELECTRA-small~\cite{electra}, achieving the best results with the DeBERTa-xsmall~\cite{deberta} model: F1, ACC >$0.9$ on validation, outperforming BB3 (F1=$0.37$, ACC=$0.76$) while being around 100x faster. Furthermore, DeBERTa-xsmall outperforms BB3 by 2 percentage points on the test dataset.

Nevertheless, despite the classifier's enhancements, we identified several issues with the usage of the User Profile (i.e. long-term memory). Predominantly, the memory was causing the Barista generative models to repetitively revert to previously discussed topics within the same session, leading to unsatisfactory outputs. Consequently, we opted not to incorporate the long-term memory module within the production version and concluded that additional work is needed to refine long-term memory integration for production-readiness. However, we hypothesize that if implemented appropriately, the integration of long-term memory could potentially enhance Barista's performance.

\subsubsection{Entity Extraction} \label{ee}
Our team extensively tested the impact of extracted entities on the quality of generated utterances during the testing of the original BB3. Although we discovered that, in some cases, entities improved the outputs of the 3B model\footnote{No improvement was discovered regarding Vicuna-based model.}, we also found that they occasionally caused repetitions. Since repetition is a significant problem identified in the BB3 model, we decided to omit this step and focus on addressing the repetition issue, as excluding the entity does not entirely solve the problem.

\subsubsection{Knowledge Extraction}
BB3's knowledge extraction task involves extracting relevant information, typically short phrases or single sentences, from a set of documents using the FiD architecture~\cite{fid}. We used the same data as Shuster~et~al.~\cite{bb3} for this task and trained two DeBERTa base models.

The first model is a classical encoder Transformer~\cite{transformer} for information extraction (question answering), using a linear layer to predict start and end token probabilities and selecting the best span accordingly.

The second model, a novel FiD-inspired architecture, processes $n$ documents and user input through a Transformer encoder, concatenates hidden states into a long-hidden state, forwards it to a Bi-LSTM~\cite{lstm} network for information fusion, and predicts start and end token probabilities (across all $n$ documents) using a linear layer.
The main advantage of such architecture is that the output probabilities are conditioned against all $n$ documents, which we hypothesize is important additional information for the model. Note that the task for the first model is to find the most likely span in a document given a user input (i.e. even for not very relevant documents, a particular span may have a high probability). The task for the second model is to find the most likely span given $n$ documents and a user input, which is an identical task to the BB3 `knowledge module', as Shuster~et~al.~\cite{bb3} calls it.


\subsubsection{Query Generation}
Based on our empirical findings, the BB3 model effectively generates high-quality search queries most of the time. However, considering its relatively high latency (see Section~\ref{deploy}): over 1.0-2.5s if all tasks are performed by the BB3 model), we sought to develop an alternative model with competitive query generation performance but a significantly reduced latency.

We designed a new model, based on FLAN-T5-base \cite{flan}, which exhibits competitive performance on our test dataset while providing over 4x speedup. Detailed results are available in Section~\ref{results}.

\subsubsection{Barista 3B Model} \label{baristaF3B}
During our testing, we discovered, besides repetition issues, a few other problems of BB3, namely: too succinct outputs, asking questions less than we would prefer, ignoring user inputs (this is mostly connected to the repetition problem), and hallucination.

We hypothesize that the fact that conversational tasks had only a minor share of the overall training dataset during training made the model less focused toward quality conversations. Furthermore, it would benefit our whole SocialBot system if our bot could control whether Barista outputs a statement or a question.
Another problem with BB3's dataset could be that the user's and BB3's utterances are separated only by a newline, which in our opinion, may confuse the model, as it may assign an utterance to the wrong side.

Therefore, we have decided to fine-tune the BB3 model with our new dataset, which consists of Topical Chat \cite{topicalChat}, Empathetic Dialogues \cite{ED}, and CCPE-M \cite{ccpe} datasets modified so that we can use control tags for generating statements, questions and other dialogue actions; and so that it separates the user's and bot's utterances with a special token. Furthermore, we created a small subset of the BB3 conversational dataset, replaced targets with parts of inputs, and used these samples as negative samples during training (i.e. maximize cross-entropy).

\subsubsection{VicuChat}
Vicuna 7B was fine-tuned on the dataset used to train the Barista 3B Model, referenced in Section~\ref{baristaF3B}, incorporating an additional prompt prior to each training sample:

\textit{You are a family friendly AI (PERSONA 1)! You are chatting with a human (PERSONA 2)! Always output just one utterance! Always end the utterance with a punctuation mark. Do not write offensive responses, you must be friendly. You don't own anything. You don't have any children nor pets! Do not switch topics too quickly! Always be polite, interesting and empathetic! Conversation:}. 

Moreover, an attempt was made to utilize LoRA  \cite{lora} to fine-tune the Vicuna model on the aforementioned data, and on a task where, given a particular context, user utterance and additional bot's utterance $bu$, an output is generated such that the subsequent bot's output could be $bu$. The same dataset as previously highlighted was used for this task, however, the final utterance became a part of a prompt, the last user's utterance was omitted, and the task consequently became the generation of the penultimate utterance. The prompt for this task is: \textit{produce the upcoming utterance, in order to ensure that '<last utterance>' can be conveyed in the succeeding turn}. Unfortunately, this endeavor proved unsuccessful as the resultant model failed to generate responses of substantial meaning.

An observation from testing revealed that the fine-tuned Vicuna model tended to output \textit{'Did you know...'}-styled trivia excessively, which can be hypothesized to stem from the Topical Chat dataset. We also noticed that certain responses by Vicuna sound artificial, a consequence which -- is once again hypothesized -- rests on the datasets used and their composition methods. Interestingly, the LoRA-trained model exhibited less susceptibility to these issues \ref{fig:ratings}.

The resultant LoRA model was designated VicuChat, maintaining its attribute as an instruction-based model. This facilitated its seamless integration into the Barista's pipeline by incorporating the query-knowledge pair in its prompt:

\textit{You are a family friendly AI (PERSONA 1)! You are chatting with a human (PERSONA 2)! Below you get additional knowledge (EXTERNAL\_KNOWLEDGE) and context of the conversation (Conversation). If the additional knowledge is relevant, generate the response based on the additional knowledge, else ignore the external knowledge. Always output just one utterance! Always end the utterance with a punctuation mark. Do not write offensive responses, you must be friendly. You don't own anything. You don't have any children nor pets! Do not switch topics too quickly! Always be polite, interesting and empathetic! EXTERNAL\_KNOWLEDGE: <QUERY-KNOWLEDGE> Conversation:}.

In order to maintain Barista reasonably fast, VicuChat uses knowledge only if required (by the Do-Search classifier), and if Evi returns an answer (Evi returns short answers, i.e. it is not necessary to use the Knowledge Extraction model). If VicuChat does not use knowledge, Barista calls VicuChat with the prompt showed in the beginning of this section.


\subsubsection{Output Decision}
We implemented several rules on when to use the Barista 3B model. If no rule is fulfilled, Barista uses VicuChat.

We present the set of rules below:
\begin{enumerate}
    \item Barista received knowledge from Evi and the user's input was a `wh-question'. (According to our empirical observation, the 3B model can generally handle this case better.)
    \item Knowledge is not included in VicuChat's response, and the estimated probability that a search is required is greater than 90\%.
    \item VicuChat's thread crashed.
    \item VicuChat returned non-sense output. (Our first version of VicuChat sometimes generated non-existing words, including characters from various alphabets, e.g. Latin, Cyrillic or Chinese. This issue was partially eliminated in the latest version.)
\end{enumerate}

The application of these governing rules results in VicuChat being deployed in about 92\% of the responses provided by Barista.

\subsection{Deployment} \label{deploy}
We implemented the core 3B model on two distinct single GPU instances -- g4dn and g5 -- to facilitate parallelization. Additionally, we set up a third instance to host the smaller task-specific models. All models were converted to the ONNX format and served through Triton and KServe servers.

VicuChat is deployed on two separate g5 instances -- one runs the primary model, and the other runs a backup to handle requests overloads.

The mean latency is 1.3~s without VicuChat and 2.2~s if VicuChat is included into the pipeline.

Furthermore, if all main and backup generative models crash, we use the Amazon ATM 20B model \cite{atm20} (using the Amazon Cobot API) as final backup.

Regarding the pipeline itself, we deployed a cascade backup system -- i.e. Alquist uses the main barista pipeline, and in case of its crash, Alquist uses a backup pipeline, which is further backed by a third pipeline.

\subsection{Results} \label{results}

Table \ref{res_table_cls} displays the performance of our classification models. The evaluation datasets, referred to as eval datasets, are obtained by merging evaluation datasets across all the data utilized for each specific task by BB3. On the other hand, our group manually curated all test datasets as described in Section~\ref{testD}. Furthermore, Table~\ref{query_gen} demonstrates the results related to query generation, and Table~\ref{knowledgeE} shows results regarding knowledge extraction. 

\begin{figure}[h!]
    \centering
    \includegraphics[width=\linewidth]{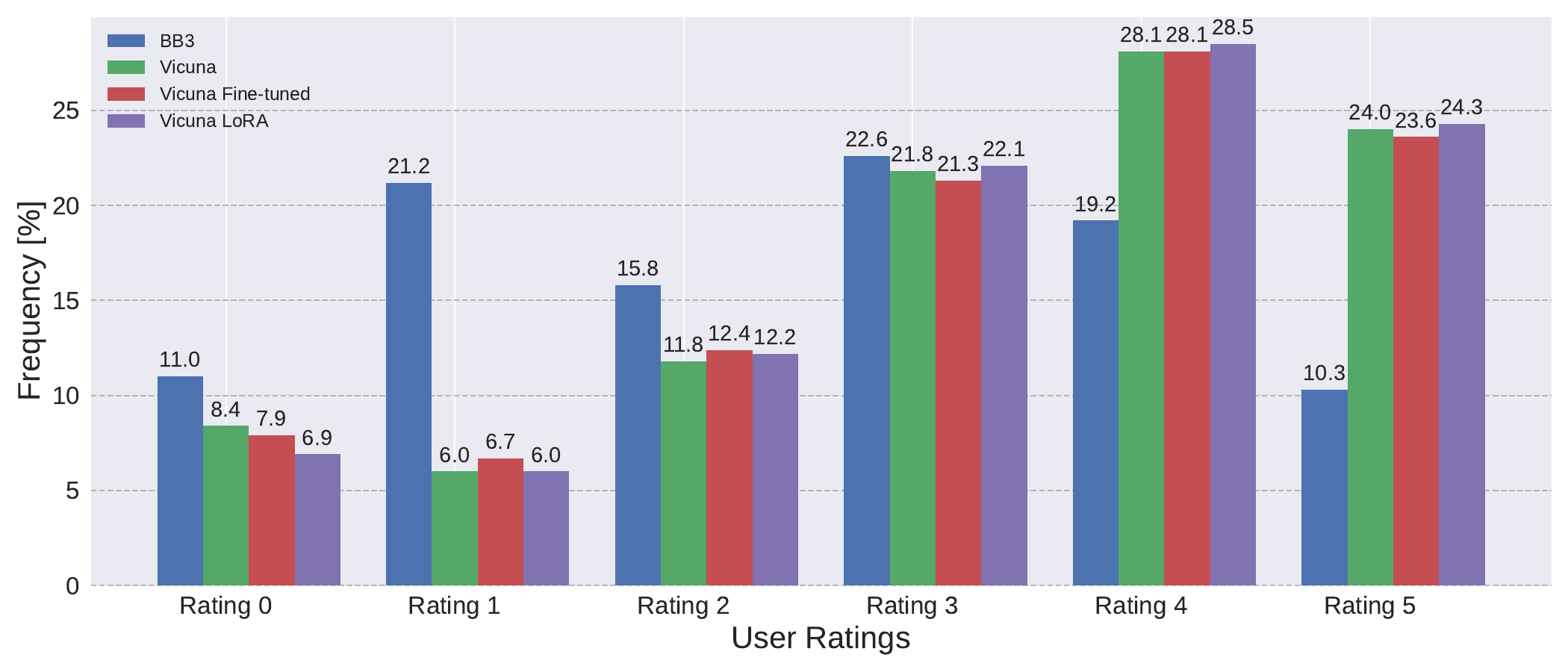}
    \caption{Test result for Barista with Vicuna trained by LoRA, fully fine-tuned, pure Vicuna and BlenderBot 3. The NRGs were rated on the scale 0-5, where 5 stands for the best rating.}
    \label{fig:ratings}
\end{figure}

To assess the performances of the original BB3 chatbot and Barista, we created a test environment using Colab\footnote{\url{https://colab.research.google.com/}}, which allowed students\footnote{We asked ten students of the Czech Technical University in Prague to participate.} to interact with either BB3 or Barista. After each response, each student was asked to rate the received utterance on a scale from 0 to 5, with 5 being the best. During the experiment, we did not reveal to the students which model they were conversing with. Our participants consisted of six students from our university, and we gathered over 800 rated turns.
Figure~\ref{fig:ratings} shows the results of Barista (BB3, pure Vicuna, fully fine-tuned VicuChat and LoRA VicuChat).

Table \ref{table:res_bb3} shows the results of our fine-tuned BB3-based 3B model (see \ref{baristaF3B}). We gathered five human annotators to converse with the original BB3 and our modified version. Each annotator had 20 conversations longer than 20 utterances. They were asked to rate the quality of conversation on a scale from 1 to 5 in several categories: overall feel, model repetition, hallucination and engagingness. From the results in Table~\ref{table:res_bb3}, we can see that the overall feel of conversation is much better than in the original BB3. Our fine-tuning of dialogue tasks and negative sampling helps in all categories besides hallucination. 

\begin{table}[t]
\centering
\small
\begin{tabular}{*{6}{c}}
 \hline
 \multirow{2}{*}{Category} & Average user rating & Average user rating\\
 & (original BB3) & (modified BB3) \\
 \hline \hline
 Overall Feel $\uparrow$ & 2.4 & \textbf{2.61}\\
 Repetition $\downarrow$ & 2.67 & \textbf{2.12}\\
 Hallucination $\downarrow$ & \textbf{2.06} & 2.10\\
 Engagingness $\uparrow$ & 2.51 & \textbf{2.93}\\
 \hline
\end{tabular}
\vspace{5pt}
\caption{Average rating over 100 conversations with the original BB3 model versus the modified BB3 (after the fine-tuning). The quality of the conversation was ranked on  scale (1-5). Depending on the rating category, an upward arrow ($\uparrow$) indicates that higher is better; a downward arrow ($\downarrow$) indicates that lower is better.}
\label{table:res_bb3}
\end{table}

\begin{table}[t]
\centering
\small
\begin{tabular}{*{6}{c}}
\hline

Model & Exact Match & F1
\\ \hline \hline

BB3 & 0.32 & 17.73\\
Deberta-base-qa & 9.45 & 26.88 \\
Deberta-large-qa & 10.11 & 27.21 \\
Deberta-large-fusion & 12.56 & 31.48\\
\hline

\end{tabular}
\caption{Knowledge extraction results. We carefully manually developed a testing dataset as described in Section~\ref{testD}.}
\label{knowledgeE}
\end{table}

\begin{table}[t]
\centering
\small
\begin{tabular}{*{7}{c}}
\hline

\multicolumn{2}{c}{Model} & \multicolumn{2}{c}{Accuracy [\%]} & \multicolumn{2}{c}{F1 [\%]} & \multicolumn{1}{c}{Speedup} \\ \hline
\multicolumn{1}{c}{task name} & \multicolumn{1}{c}{model name} & \multicolumn{1}{c}{Eval} & \multicolumn{1}{c}{Test} & \multicolumn{1}{c}{Eval} & \multicolumn{1}{c}{Test} & \multicolumn{1}{c}{Test + Eval}
\\ \hline \hline

do search? & BB3 & 84.0 & 62.9 & 86.2 & 71.2 & 1x \\ 
do search? & ELECTRA-small & 88.1 & 69.8 & 89.3 & 73.5 & 50x \\
do search? & BERT-tiny & 90.3 & 78.1 & 90.6 & 76.4 & \textbf{192x}\\
\textbf{do search?} & \textbf{DeBERTa-xsmall} & \textbf{92.2} & \textbf{84.9} & \textbf{92.3} & \textbf{82.5} & 17x\\
\hline
access user profile? & BB3 & 76.1 & 69.5 & 36.9 & 69.5 & 1x\\ 
access user profile? & ELECTRA-small & 87.3 & 59.8 & 42.5 & 59.8 & \textbf{50x}\\
\textbf{access user profile?} & \textbf{DeBERTa-xsmall} & \textbf{92.1} & \textbf{71.4} & \textbf{92.1} & \textbf{71.4} & 17x\\ \hline

\end{tabular}
\caption{Performance of BB3 model and our fine-tuned models on two classification tasks: do-search? and access-user-profile?}
\label{res_table_cls}
\end{table} 

\begin{table}[ht]
\centering
\small
\begin{tabular}{*{6}{c}}
\hline

Model & Accuracy & Weak Accuracy & SacreBLEU (EVAL) & Speed up
\\ \hline \hline

BB3 & \textbf{73.6} & 96.2 & 12.1 & 1x\\
FLAN-Alpaca-large \cite{alpaca} & 56.6 & \textbf{98.1} & \textbf{13.6} & 2.2x\\
\textbf{FLAN-T5-base} & 54.7 & \textbf{98.1} & 12.2 & \textbf{4.3x}\\
\hline

\end{tabular}
\caption{We evaluated the performance of the BB3 model and our fine-tuned models in query generation using a test dataset. We assessed the results based on three criteria: exact match, strict semantic match (in which the generated query is not an exact match but holds the same semantic meaning as the label), and relevance (where the generated query pertains to the topic but may not be as precise). It is important to note that a query with an exact match or strict semantic match is also considered relevant. The accuracy metric is a combination of exact match and strict semantic match, while weak accuracy corresponds to relevance.}
\label{query_gen}
\end{table} 

\section{APIHub}
\label{apihub}
When Barista's Do Search Classifier outputs a positive prediction, a request is triggered to APIHub, our search API. A single call to APIHub returns a set of documents from 4 different services (ordered by usefulness):

\begin{itemize}
    \item Amazon Evi -- a question answering system
    \item DuckDuckGo\footnote{\url{https://duckduckgo.com}} (DDG) -- an internet search engine
    \item Wikipedia\footnote{\url{https://dumps.wikimedia.org}} Introductions -- our database of lead sections of Wikipedia articles
    \item News Retrieval API -- a service included in the Cobot Toolkit
\end{itemize}

Within the request, we specify two types of queries -- one generated by our query generator and a \emph{keyword query}. \emph{Keyword query} is created using SpaCy\footnote{\url{https://spacy.io}} by extracting nouns and numbers from the original query. The \emph{keyword query} is used for the Wikipedia introductions and News Retrieval API service since they require a keyword, whereas Evi and DuckDuckGo work best with a longer query.

The core APIHub service is Amazon Evi. It is a question-answering system which can respond in a natural language to its queries. We found that Evi's high-quality answers are the best knowledge source for Barista.
Second, we use the internet search engine DDG. After searching the internet, we return the content of the top-3 found web pages.
We have also created our database of Wikipedia articles. In order to simplify the work of Barista's Knowledge Extraction module and not flood it with too many documents, we have retained only the introductory sections of Wikipedia articles, which summarize the whole article and include the key knowledge to be extracted.
The last service is the News Retrieval API from the Cobot Toolkit, which was provided for the competition. When provided a query, it returns summarized recent news articles. This allows Barista to be informed about recent events the user may mention.

In order to sustain a large number of requests, APIHub works asynchronously by using the FastAPI framework\footnote{\url{https://fastapi.tiangolo.com/}}. Also, we have set a timeout of 1.5~s and included several optimizations to speed up the whole process.
The first optimization is to not wait for the output of DDG if Evi provides an answer. Since DDG is APIHub's slowest service, we limit its use as much as possible. However, Evi does not always have an answer to the query, so DDG's slowness is compensated by its ability to always find a relevant set of documents.
Another optimization is that during DDG's search, we ignore results from Wikipedia since we can expect them to be present in the Wikipedia Introductions service.
Finally, we cache APIHub results for two weeks using an Amazon ElastiCache Redis instance\footnote{\url{https://aws.amazon.com/elasticache/redis/}}. With a cache hit rate of 50-60\%, we answer a significant part of APIHub's queries instantaneously.

\section{Safety Framework}
\subsection{Introduction}
In recent years, the development of neural conversational models has led to significant progress in the field of natural language processing, particularly in the area of generative open-domain chatbots \cite{zhang2019dialogpt,adiwardana2020meena, roller2020blenderbot}. However, the widespread deployment of these models in real-world scenarios has been hindered by concerns regarding dialogue safety \cite{micro2016exam}. While the use of transformer-based language models pretrained on large-scale corpora has enabled the creation of increasingly sophisticated chatbots, these models often produce uncontrollable and unpredictable outputs that can result in offensive, toxic, or otherwise unsafe language.

In this section, we analyze and demonstrate the existence of significant safety problems in existing dialogue systems across multiple datasets for multiple language models. Building on the analysis results, we propose a potential solution to this problem in the form of multiple classifiers with a combination of a rule-based system. This straightforward solution could effectively enhance the safety of chatbots by identifying and filtering out potentially unsafe responses.

\subsection{Related work}
The issue of dialogue safety in chatbots has attracted significant research interest in recent years. A number of studies have focused on identifying and mitigating the risks associated with unsafe language generation in conversational models~\cite{herel2022preserving, political2021Liu, hateoffensive}. For instance, the Wikipedia Toxic Comments dataset \cite{Wulczyn2017WTCDataset} with 100k human-labelled data and Twitter dataset \cite{hateoffensive} with human-annotated 240k hate speech and offensive language tweets.
Previous research also explored other dialogue safety topics such as political prudence \cite{bang2021assessing}, stereotype \cite{barikeri2021redditbias}, and many debiasing methods for certain domains were proposed \cite{Zhao2018LearningGW, Rudinger2018GenderBI, political2021Liu}.

\subsection{Experiments}
The first goal of our experiment was to determine the extent to which chatbots can produce unsafe content in their responses. We trained classifiers on several hate, offensive, and biased datasets \cite{cyberbully, hateoffensive, sun-etal-2022-safety, stereoset}. We used these classifiers to identify unsafe content in the responses generated by chatbots.

We used ChatGPT as a badly behaved user and generated responses that could potentially contain unsafe content. We then let ChatGPT converse with state-of-the-art dialogue language models for 1000 turns. During this interaction, we measured the percentage of responses generated by the language models that were flagged by our classifiers as containing unsafe content. We have also used human evaluators to judge how many flagged responses were false positives.

\subsection{Results}
\begin{figure}[h!]
\centering
\includegraphics[width=\linewidth]{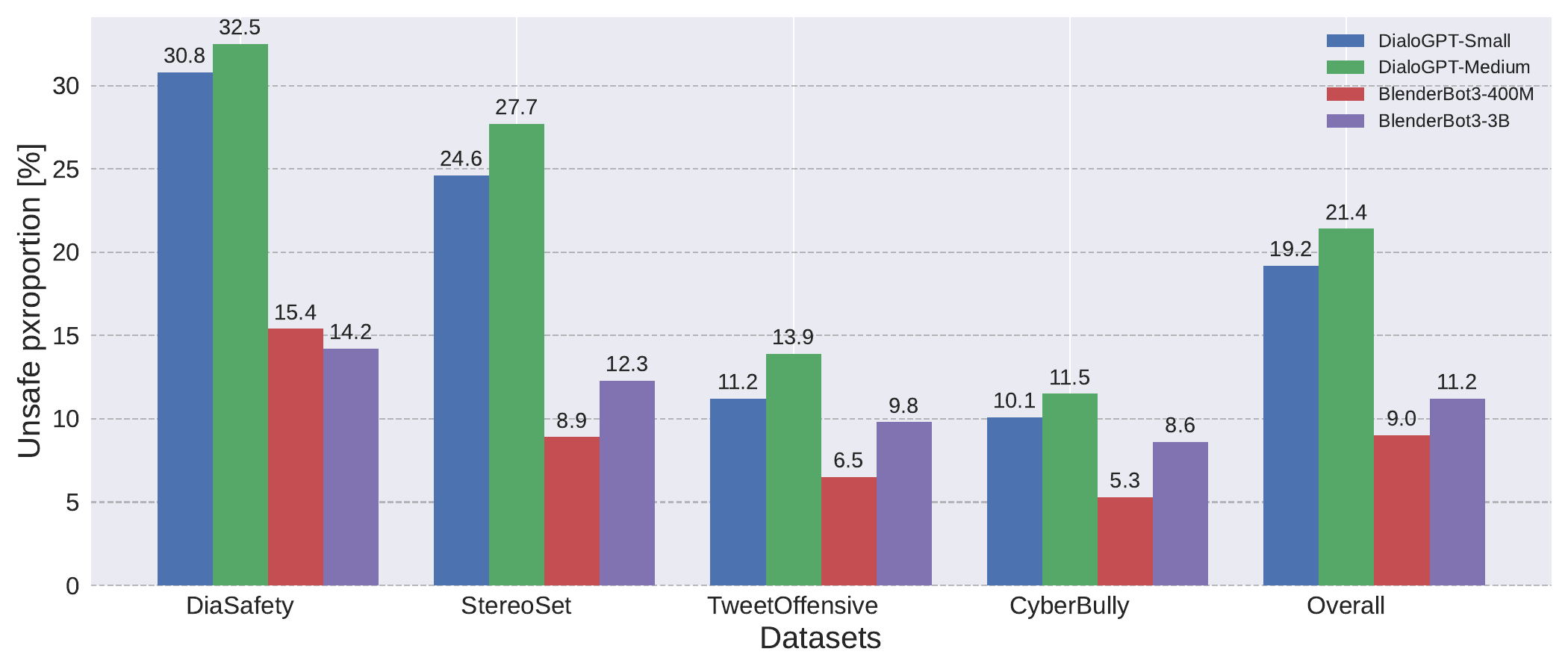}
\caption{Results show unsafe content as an Unsafe proportion in \% (lower is better) across four datasets - DiaSafety, StereoSet, TweetOffensive, CyberBully for each language model: DialoGPT-Small, DialoGPT-Medium, BlenderBot3-400M, BlenderBot3-3B. The "Overall" was computed as an average of values for a selected language model.}
\label{fig:results}
\end{figure}
Our study found that a considerable proportion of the responses produced by the dialogue language models were identified as containing potentially unsafe content. Notably, as depicted in Figure \ref{fig:results}, our experiments suggest that larger models do not necessarily result in safer responses. These findings align with DiaSafety results \cite{sun-etal-2022-safety}. One possible explanation is that smaller models tend to produce more generic responses, which are less prone to generating unsafe or biased language.

Upon conducting a manual examination of the flagged responses, we observed that, on average, 32\% of the flagged responses were identified as false positives. This finding implies that the current datasets used for training the classifiers are not entirely representative of the nuances of harmful language. Hence, we propose supplementing them with rule-based systems to achieve greater accuracy. We expand on our proposed approach in the next section.

\subsection{Our method: Combined Safety Filter}
\label{section:combined}
Experiment findings imply that the current datasets used for training the classifiers are not entirely representative of the nuances of harmful language. Hence, we propose supplementing them with rule-based systems to achieve greater accuracy.

\begin{figure}[ht]
\centering
\includegraphics[width=0.4\linewidth]{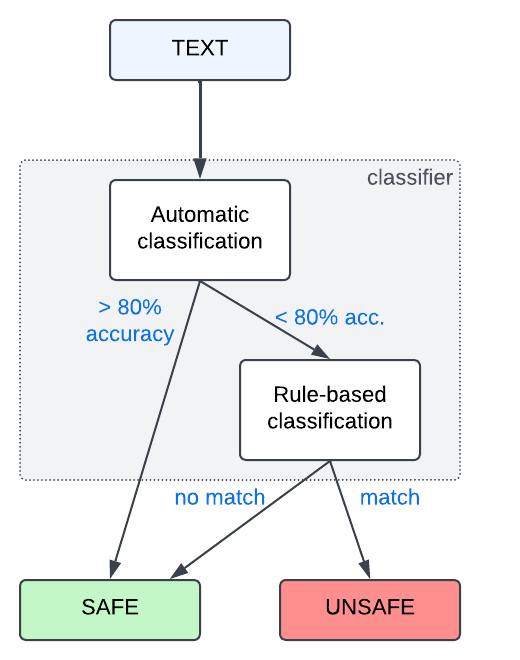}
\vspace{-10pt}
\caption{Visualization of combined approach.}
\label{fig:concept}
\end{figure}

Our algorithm, illustrated in Figure \ref{fig:concept}, utilizes both fastText classifiers \cite{fasttext} (\autoref{appendix:ft-hyperparameters} describes used hyperparameters) and a rule-based string-matching system to classify input sentences. The rule-based system consists of a list of unsafe words and phrases, and if there is a match between the input text and the unsafe list, the input is marked as unsafe.

In our combined approach, the text is classified as safe if it is flagged as such by all classifiers, with an average accuracy of over 80\%. If the accuracy is below 80\%, the rule-based system determines whether the input should be flagged as unsafe. This concept is depicted in Figure \ref{fig:concept}.

As shown in Table \ref{table:res}, the combination of automatic classification and rule-based system achieves a 6.5\% higher F1 score compared to using only the fastText classifiers. These results support the effectiveness of our proposed approach in improving the accuracy of harmful language detection.

\begin{table}[h!]
\centering
\small
\begin{tabular*}{0.52\linewidth}{l|p{2cm}|p{2cm}}
 \toprule
 \multirow{2}{*}{Dataset} & Automatic & Combined \\
 & F1 score $\uparrow$ & F1 score $\uparrow$\\
 \midrule
 CyberBully \cite{cyberbully} & 0.827 & \textbf{0.901}\\
 DiaSafety \cite{sun-etal-2022-safety}& 0.698 & \textbf{0.786}\\
 TweetOffensive \cite{hateoffensive}& 0.893 & \textbf{0.923} \\
 StereoSet \cite{stereoset}& 0.678 & \textbf{0.745} \\
 \bottomrule
\end{tabular*}
\vspace{5pt}
\caption{Results of automatic versus combined approach on F1 score. An upward arrow ($\uparrow$) indicates that higher is better.}
\label{table:res}
\end{table}

\subsection{Conclusion}
In conclusion, we address the critical issue of dialogue safety in chatbots and highlight the significant safety problems that exist in current conversational models. The findings demonstrate the urgent need for more effective safety measures to be incorporated into dialogue systems to ensure their safe deployment in real-world applications. By proposing a solution that combines multiple classifiers with a rule-based system, we offer a promising approach to improving the dialogue safety of chatbots.

\section{Dialogue Management and Large Language Models}
\label{section:DM_LLMs}
The dialogue manager of the system utilizes Dialogue Selector and Dialogue Trees proposed by Konrád~et~al.~\cite{konrad2021alquist}. However, the development of Barista opened a new problem of how to incorporate large language models (LLMs) into dialogue management. While human-designed dialogue trees offer engaging conversations meticulously crafted by dialogue designers, they lack the flexibility of LLMs to respond to unforeseen inputs.

To enhance the system's abilities, we propose four approaches for incorporating generative models into dialogue management: Handling of Out-of-domain (OOD) Inputs, Handling of Proactive Questions, Hybrid Dialogues, and Inserting LLMs into Dialogue Trees. The LLM loop is central to all four strategies, managing the conversational flow for several turns.

\subsection{Structure of the LLM Loop}
\begin{figure}[h!]
    \centering
    \includegraphics[width=0.18\textwidth]{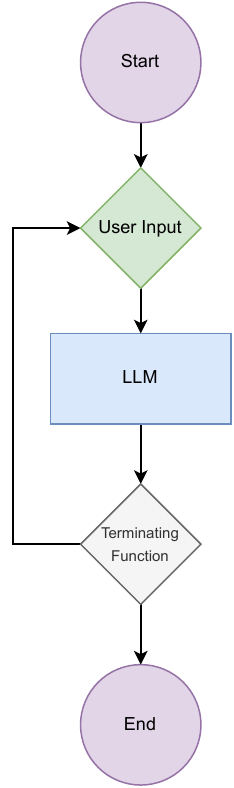}
    \caption{LLM loop}
    \label{fig:llm_loop}
\end{figure}

The LLM loop comprises the user input, the generated response, and a terminating function, the latter of which determines whether the loop continues or ends (see Figure~\ref{fig:llm_loop}). The terminating function operates with three criteria: 

\begin{itemize}
    \item \textbf{Turn Counting:} It keeps track of the number of turns handled by the LLM. If this count exceeds a predefined threshold, the loop is terminated to prevent the situation in which the topic of conversation wanders off the original topic. Also, the probability of a nonsensical response increases with the number of generated responses in a row. Thus, the turn counting mechanism ensures that the conversation remains coherent and does not deviate from the topic.

    \item \textbf{Regex Matching:} The function detects phrases such as ``Alright'', ``I don't know'' or ``I want to chat about something else'', which suggest user disinterest in the current topic of conversation. If any of these phrases or their variations are detected, the loop is terminated to allow for a change in conversational direction and maintain user interest.

    \item \textbf{ODES Classifier~\cite{le2023improving}:} The class predicted by the ODES classifier serves as another criterion for terminating the LLM loop. If the classifier predicts one of the classes \emph{USER\_DISINTEREST}, \emph{USER\_INITIATED\_TOPIC\_SWITCH} or \emph{USER\_REQUEST\_STOP} indicating that the conversation is no longer engaging to the user, the loop is terminated. This ensures that the SocialBot's responses remain engaging and adaptive to user preferences.
\end{itemize}

\subsection{Integration of LLMs into Conversation}

The proposed approaches for incorporating generative models into dialogue management (see Figure~\ref{fig:llm_loopUtilization}) use the LLM loop and aim to expand the flexibility and adaptability of Alquist 5.0, providing a more engaging and robust user experience. The approaches are:

\begin{figure}[htb]
     \centering
     \begin{subfigure}{0.65\textwidth}
         \includegraphics[width=\textwidth]{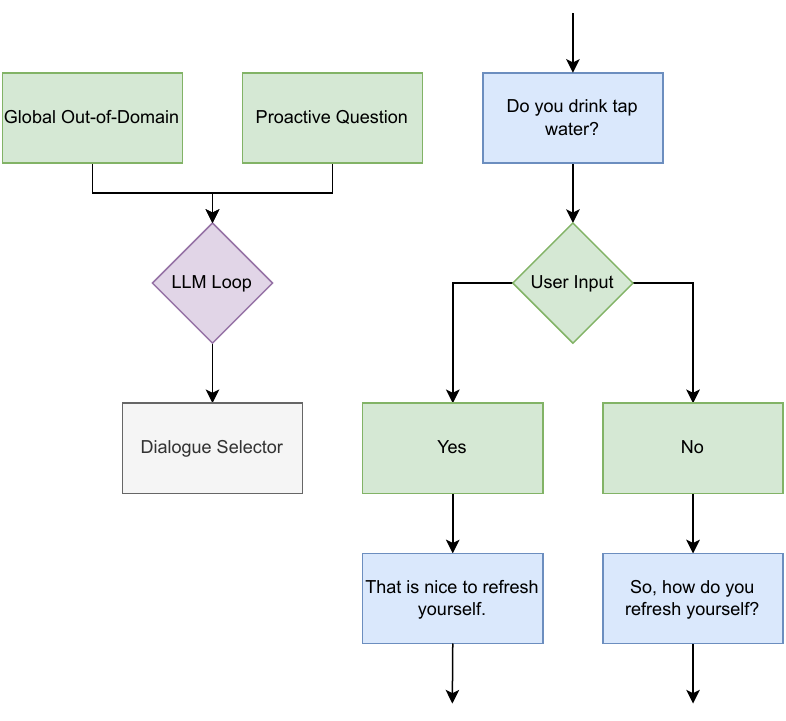}
         \caption{Handling of Out-of-Domain and Proactive Questions}
         \label{fig:llm_ood}
     \end{subfigure}
     \hfil
     \begin{subfigure}{0.16\textwidth}
         \includegraphics[width=\textwidth]{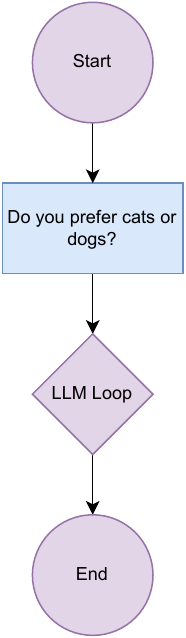}
         \caption{Hybrid Dialogues}
         \label{fig:llm_hybridDialogue}
     \end{subfigure}
     \medskip
     \begin{subfigure}{0.65\textwidth}
         \includegraphics[width=\textwidth]{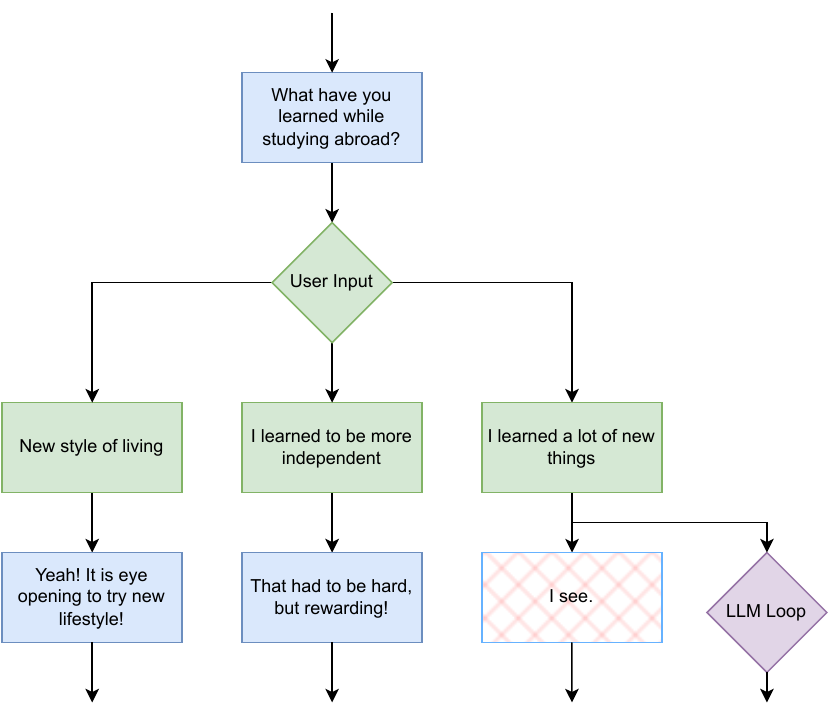}
         \caption{Inserting LLMs into Dialogue Trees (LLM Loop replaces non-informative acknowledgement ``I see'')}
         \label{fig:llm_tree}
     \end{subfigure}
        \caption{Approaches for incorporating generative models into dialogue management. Green nodes represent intents recognized in users' inputs. Blue nodes represent SocialBot's responses.}
        \label{fig:llm_loopUtilization}
\end{figure}

\begin{enumerate}
    \item \textbf{Handling of out-of-domain (OOD):} Konrád~et~al.~\cite{konrad2021alquist} proposed using the LLM to handle unexpected or unrecognized user inputs. By leveraging OOD detection techniques, the system determines whether the user input falls outside the scope of the current intents that can handle the continuation of the conversation. If an OOD input is identified, control over the conversation's flow transfers to the LLM, which generates a response to the unexpected input and Dialogue Selector selects the next dialogue tree to execute. Konrád~et~al.~\cite{konrad2021alquist} proposed to use DialoGPT \cite{zhang2019dialogpt}. In our work, we propose to innovate the approach by using Barista. Additionally, as opposed to Konrád~et~al.~\cite{konrad2021alquist}, we propose generating more than one response. Instead, we generate responses for a predetermined number of turns through the LLM loop (see \autoref{fig:llm_ood}). This approach enables SocialBot to handle unforeseen user inputs globally. Moreover, the LLM loop deepens the conversation toward an out-of-domain topic.

    \item \textbf{Handling of Proactive Questions:} Proactive questions, i.e. questions asked by the user, represent a unique challenge. Dialogue trees tend to ignore proactive questions by default because they are primarily designed to recognize user answers and proceed accordingly without considering additional questions directed toward the SocialBot. For instance, when a bot asks, ``What is your favorite ice cream flavor?'' and the user responds, ``Vanilla. What's yours?'', the intent recognition component recognizes intent `Vanilla', for which the system may only provide information about the popularity of vanilla, disregarding the user's reciprocal question. The out-of-domain intent, possibly helpful in this situation, is not triggered because there is some amount of semantic similarity between user input and recognized intent. In contrast, LLMs can effectively address this issue, displaying greater flexibility in handling such questions. By utilizing the Cobot Toolkit Punctuation API for properly punctuating user inputs and subsequently identifying question marks, the system can redirect the conversational flow to the LLM loop (see Figure~\ref{fig:llm_ood}). The LLM loop responds appropriately to the user's proactive question while delving deeper into the relevant topic for the predetermined number of dialogue turns. This approach allows SocialBot to respond to proactive questions making it more coherent.

    \item \textbf{Hybrid Dialogues:} This approach enriches the dialogue system by merging the advantages of both human-designed dialogue trees and LLM-generated responses. Hybrid dialogues start with an engaging question, fun fact or comment authored by a dialogue designer to spark user interest and set the context for the conversation. When the user responds to this initial question, the LLM loop is triggered for a predetermined number of turns. The LLM can then provide contextually appropriate responses to user inputs, dynamically adapting the conversation to the user's interests and preferences (see Figure~\ref{fig:llm_hybridDialogue}). This approach allows for the rapid development of engaging conversations, combining the strengths of structured dialogue trees and the flexible responses generated by large language models.

    \item \textbf{Inserting LLMs into Dialogue Trees:} To improve the responsiveness and depth of interactions within existing dialogue trees, we pinpointed several intents followed by non-informative acknowledgements, such as ``I see'', ``Got it'' or ``I understand'' typically handling user responses to open-ended questions seeking opinions. By replacing these brief and non-informative acknowledgements with the LLM loop (see Figure~\ref{fig:llm_tree}), the SocialBot gains the ability to provide more contextually relevant, coherent, and engaging reactions to users' opinions. The LLM loop dynamically directs the conversation according to user input, enabling deeper interactions that can span multiple conversational turns and allowing users to express their thoughts more freely and expand on them.
\end{enumerate}

These four proposed LLM integration approaches in a SocialBot's dialogue management enhance its ability to adapt to unexpected inputs and proactive questions, offer a more comprehensive range of engaging content, and provide more natural and dynamic responses to user inputs. Although we employ Barista as the LLM, all the approaches are independent of any specific LLM architecture. Thus, the proposed strategies have broad applicability to many conversational AI systems.

\section{User Interface}
\label{section:UI}

This year's SocialBot Grand Challenge is different from previous edition in the fact that for the first time, the SocialBots must be able to support multimodal devices. Therefore, we also focused on the visual aspect of our bot. In the following sections, we will describe the supported APL\footnote{\url{https://developer.amazon.com/en-US/docs/alexa/alexa-presentation-language/add-visuals-and-audio-to-your-skill.html}} templates, and the three key features of our templates: Karaoke mode, Preserve mode, and 3D Persona background.

\subsection{Supported APL Templates}
\begin{figure}[ht]
     \centering
     \begin{subfigure}{0.49\textwidth}
         \includegraphics[width=\textwidth]{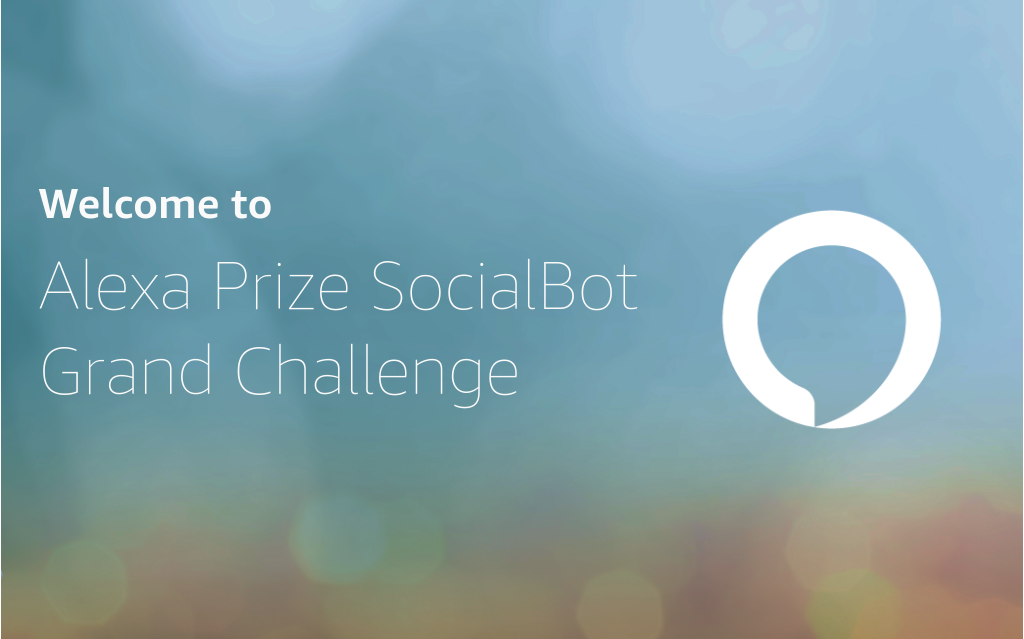}
         \caption{SocialBot Landing}
         \label{fig:socialbot_landing}
     \end{subfigure}
     \hfil
     \begin{subfigure}{0.49\textwidth}
         \includegraphics[width=\textwidth]{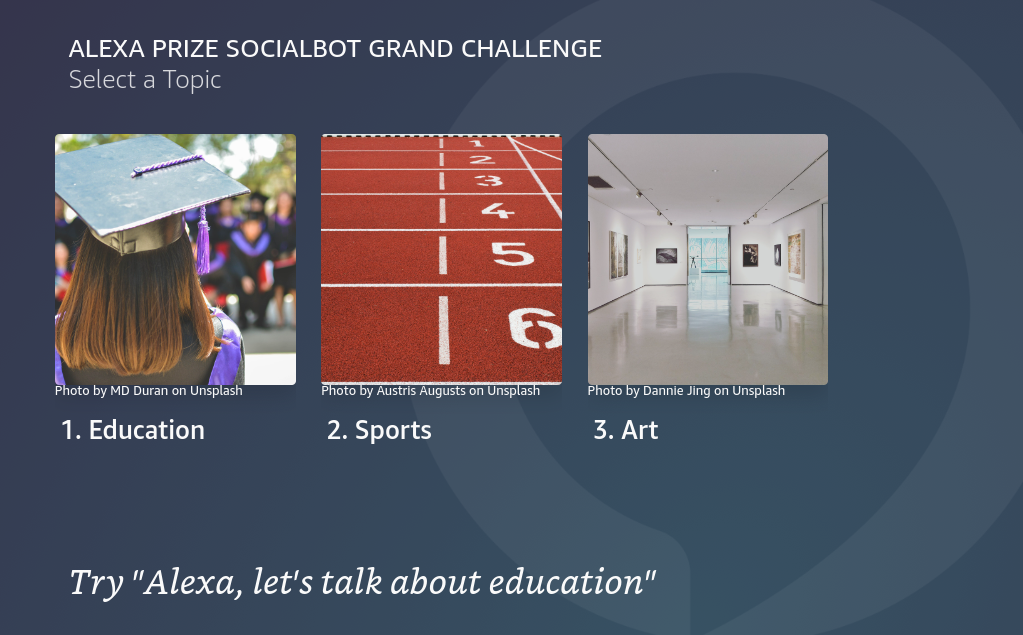}
         \caption{SocialBot Image List}
         \label{fig:socialbot_img_list}
     \end{subfigure}
     \medskip
     \begin{subfigure}{0.49\textwidth}
         \includegraphics[width=\textwidth]{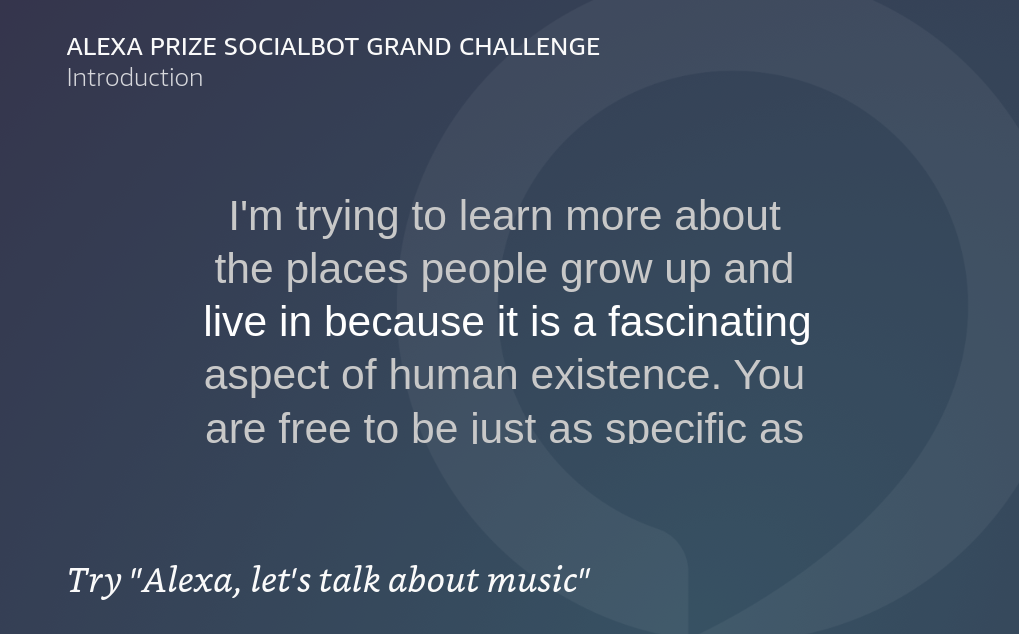}
         \caption{SocialBot Karaoke Chat}
         \label{fig:socialbot_karaoke_chat}
    \end{subfigure}
    \hfil
     \begin{subfigure}{0.49\textwidth}
         \includegraphics[width=\textwidth]{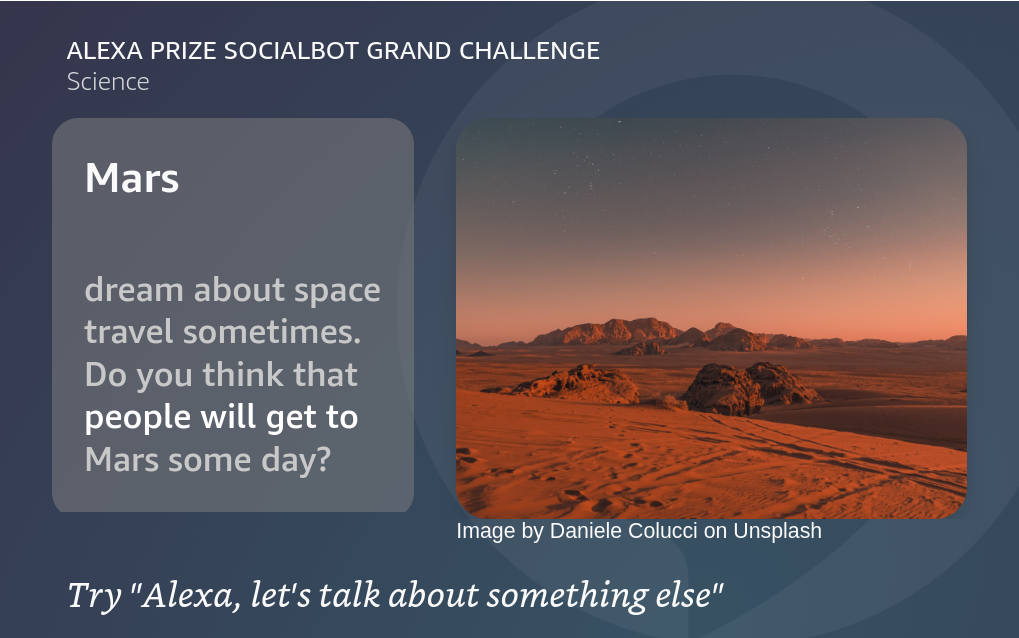}
         \caption{SocialBot Karaoke Detail}
         \label{fig:socialbot_karaoke_detail}
     \end{subfigure}
     \medskip
     \begin{subfigure}{0.62\textwidth}
         \centering
         \includegraphics[width=\textwidth]{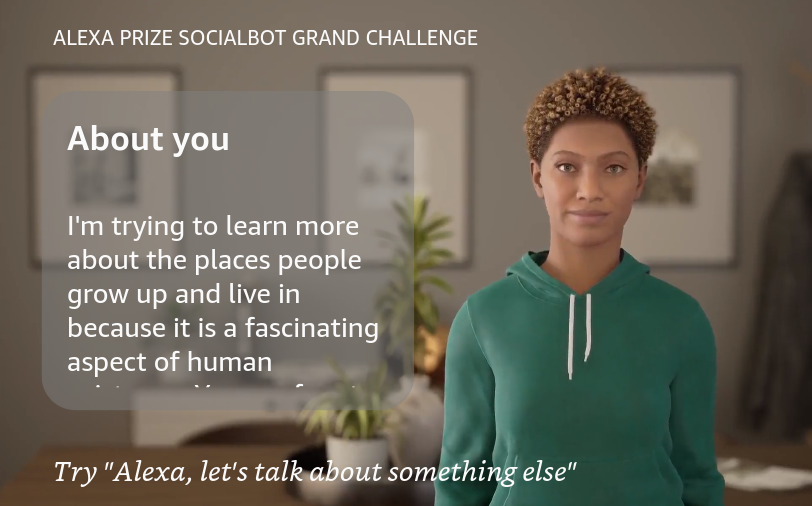}
         \caption{SocialBot Karaoke Avatar on a horizontal device}
         \label{fig:socialbot_karaoke_avatar}
     \end{subfigure}
     \hfil
     \begin{subfigure}{0.37\textwidth}
         \includegraphics[width=\textwidth]{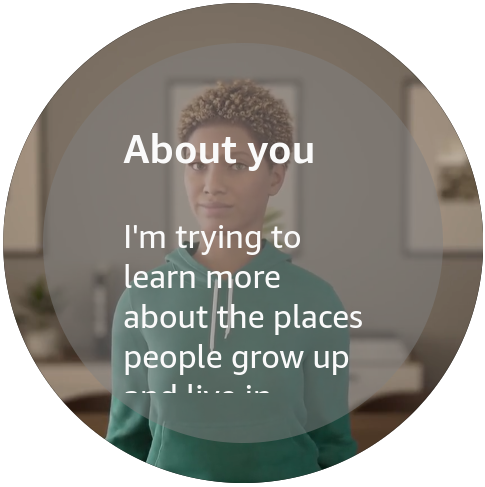}
         \caption{SocialBot Karaoke Avatar on a round small device}
         \label{fig:socialbot_karaoke_detail_round_small}
     \end{subfigure}
        \caption{Utilized APL templates}
        \label{fig:used_apl_templates}
\end{figure}

We employ five different APL templates in our SocialBot. Specifically, we are utilizing the SocialBot Landing, SocialBot Image List, SocialBot Karaoke Chat, SocialBot Karaoke Detail, and SocialBot Karaoke Avatar templates. See Figure \ref{fig:used_apl_templates} for a showcase of the utilized APL templates in our bot. The first two APL templates are available in the Cobot Toolkit\footnote{\url{https://www.amazon.science/blog/ai-tools-let-alexa-prize-participants-focus-on-science}}. The last three templates were implemented by us.

\paragraph{SocialBot Landing} SocialBot Landing template is used as the first screen to welcome the users (see Figure~\ref{fig:socialbot_landing}). 

\paragraph{SocialBot Image List} SocialBot Image List template is used for switching topics, either when a user asks to change a topic or when a topic has come to its natural end (see Figure~\ref{fig:socialbot_img_list}).

\paragraph{SocialBot Karaoke Chat} We use this template whenever we cannot use the SocialBot Karaoke Avatar template (see Figure \ref{fig:socialbot_karaoke_chat}). We created this template based on the SocialBot Chat template from the Cobot Toolkit.

\paragraph{SocialBot Karaoke Detail} SocialBot Karaoke Detail template is used in our scripted subdialogues (see Figure \ref{fig:socialbot_karaoke_detail}). We created this template based on the SocialBot Detail template from the Cobot Toolkit.

\paragraph{SocialBot Karaoke Avatar} We use SocialBot Karaoke Avatar template everywhere apart from the initial introduction screen, scripted subdialogues and change of topic (e.g. in the dialogues about the user name, how user is doing, etc., see Figure \ref{fig:socialbot_karaoke_avatar}).

\subsection{Karaoke Mode}

By karaoke mode, we mean the synchronization of the voice with the text displayed on the screen\footnote{\url{https://developer.amazon.com/en-US/docs/alexa/alexa-presentation-language/apl-speech-and-text-synchronization-for-text-blocks.html}}. In karaoke mode, the spoken text is highlighted. In order to utilize this feature, it is necessary to provide speech data in plain text or by utilizing \textit{Speech Synthesis Markup Language (SSML)}\footnote{\url{https://developer.amazon.com/en-US/docs/alexa/custom-skills/speech-synthesis-markup-language-ssml-reference.html}} expressions. Before an Alexa-enabled device consumes this data, the data needs to be converted into spoken words. One can employ the \textit{ssmlToSpeech transformer}\footnote{\url{https://developer.amazon.com/en-US/docs/alexa/alexa-presentation-language/apl-transformers.html\#ssmltospeech-transformer}} to facilitate this conversion process, which converts the text to speech and removes SSML tags from the SSML expression. Consequently, when the response is being spoken, the device will automatically scroll to display the portions of text currently being read. To have the response read aloud, one must employ \textit{Alexa.Presentation.APL.ExecuteCommands}\footnote{\url{https://developer.amazon.com/en-US/docs/alexa/alexa-presentation-language/apl-interface.html\#executecommands-directive}} directive and utilize the \textit{SpeakItem} command.

\subsection{Preserve Mode}
We developed a preserve mode for a more straightforward configuration of APL templates. This mode allows us to specify that the chosen APL template, alongside the set parameters, will be remembered until another template is specified. This template will be facilitated even when no APL template is configured until another template is employed with the preserve mode.

\subsection{3D Persona Background}
When talking with a SocialBot, the users might imagine the bot in a human body. To create a more positive, realistic, and pleasant user experience, we have created a 3D Persona -- Alquistyna -- representing our bot using MetaHuman\footnote{\url{https://www.unrealengine.com/en-US/metahuman}}. In order to employ Alquistyna, we first rendered it in a friendly environment using Unreal Engine~5\footnote{\url{https://www.unrealengine.com/en-US/unreal-engine-5}}. After that, we generated a short animated sequence played in an infinitive loop as background on the screen. To ensure low latency, we compressed the video sequence to reduce the size of the videos. In total, we generated nine different animations, each in two settings (one with Alquistyna standing in the middle, which is used on vertical and square devices, as shown in Figure \ref{fig:socialbot_karaoke_detail_round_small}, the other with Alquistyna standing more to the right, which is used on horizontal devices, as shown in Figure \ref{fig:socialbot_karaoke_avatar}). We generated the two settings to ensure an excellent user experience across all types of Alexa devices. Furthermore, we also tweaked the responsive rules to even more improve the user experience. In the following section, we will briefly describe all nine animations.

\subsubsection{3D Persona Background Types}

\begin{figure}[h!]
     \centering
     \begin{subfigure}{0.49\textwidth}
         \centering
         \includegraphics[width=\textwidth]{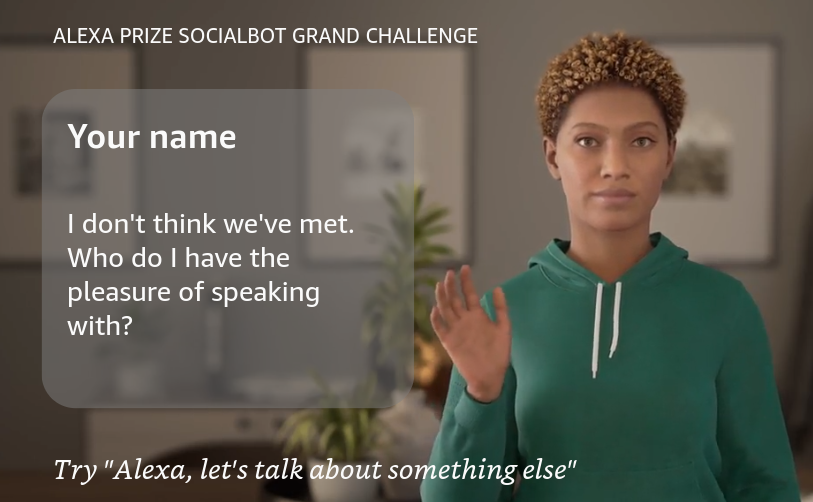}
         \caption{SocialBot Karaoke Avatar Greetings}
         \label{fig:avatar_wave}
     \end{subfigure}
     \hfil
     \begin{subfigure}{0.49\textwidth}
         \includegraphics[width=\textwidth]{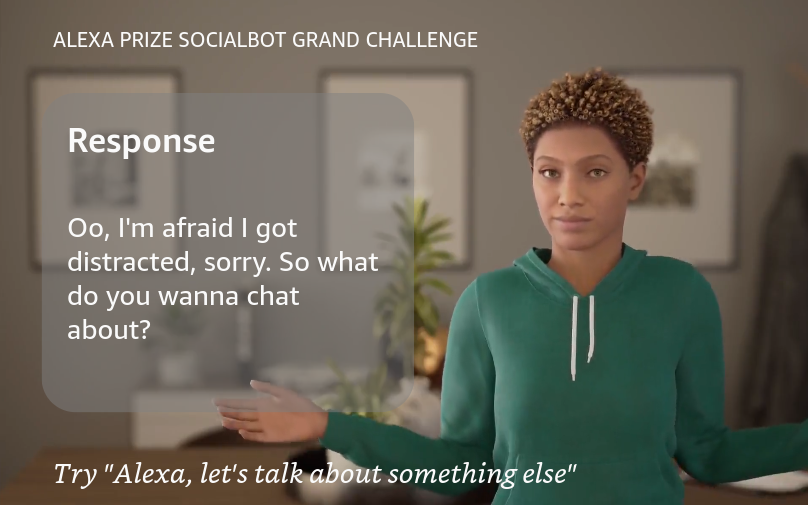}
         \caption{SocialBot Karaoke Avatar Error}
         \label{fig:avatar_error}
     \end{subfigure}
        \caption{General SocialBot Karaoke Avatar showcase}
        \label{fig:general_avatars}
\end{figure}

\paragraph{General} The following three animations are general and not thematic.

\textbf{Idle} -- The most generic animation, in which Alquistyna is looking at the user and smiles (see Figure \ref{fig:socialbot_karaoke_avatar}).

\textbf{Greetings} -- In this animation, Alquistyna waves at the user to greet them (see Figure~\ref{fig:avatar_wave}).

\textbf{Error} -- In this animation, Alquistyna shrugs her shoulders. We use this animation when there was some internal error and we could not render a response (see Figure~\ref{fig:avatar_error}).

\begin{figure}[h!]
    \centering
    \includegraphics[width=0.95\textwidth]{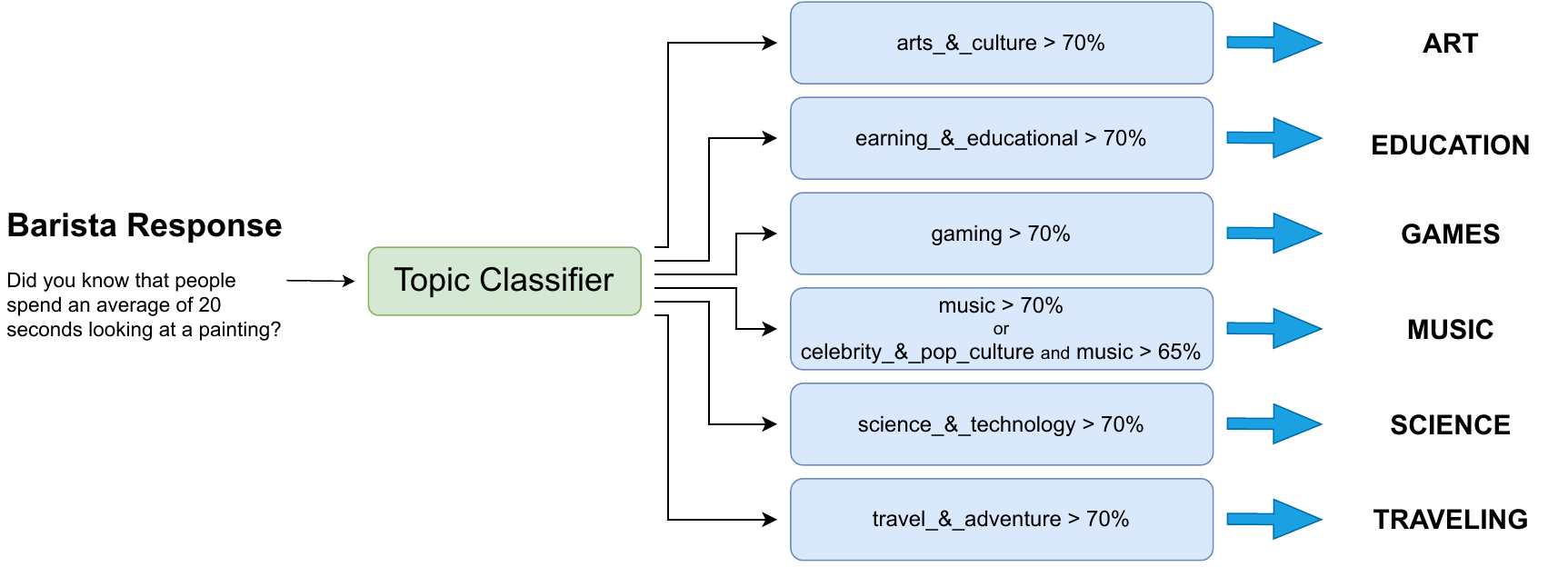}
    \caption{A schema of topic classification.}
    \label{fig:topic_classification}
\end{figure}

\paragraph{Topic specific} For the visuals to be more relevant to the current topic, we recognized six topics (Gaming, Education, Music, Art, Traveling, and Science) for which we created a specific, more relevant animation. We took the general idle animation and put Alquistyna into a thematic environment. We employed these thematic animations in our hybrid scripted dialogues and unscripted conversation. We assigned to each hybrid scripted dialogue one animation. Whenever the conversation becomes unscripted (e.g. because out-of-domain was detected), we use a topic classifier model~\cite{antypas-etal-2022-twitter} to classify the topic based on the response of our Barista NRG model and based on the detected topic we set the relevant animation. Figure~\ref{fig:topic_classification} shows the flow of the topic classification.

\begin{figure}[h!]
     \centering
     \begin{subfigure}{0.49\textwidth}
         \includegraphics[width=\textwidth]{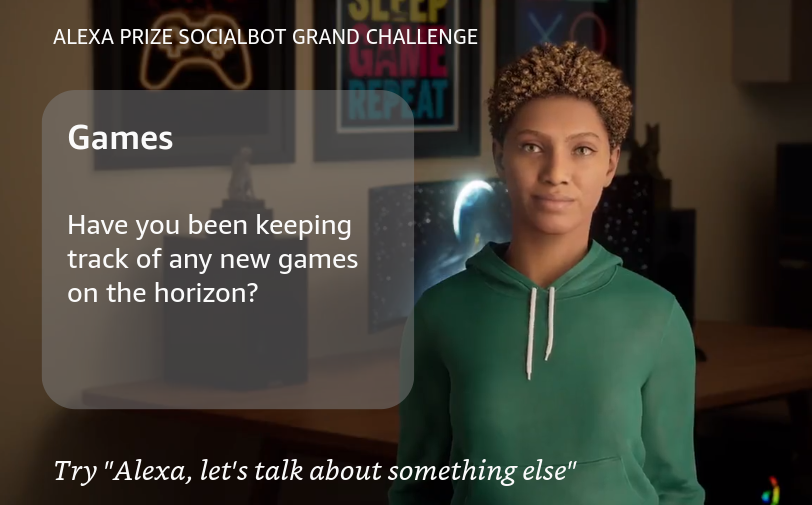}
         \caption{SocialBot Karaoke Avatar Gaming}
         \label{fig:avatar_games}
     \end{subfigure}
     \hfil
     \begin{subfigure}{0.49\textwidth}
         \includegraphics[width=\textwidth]{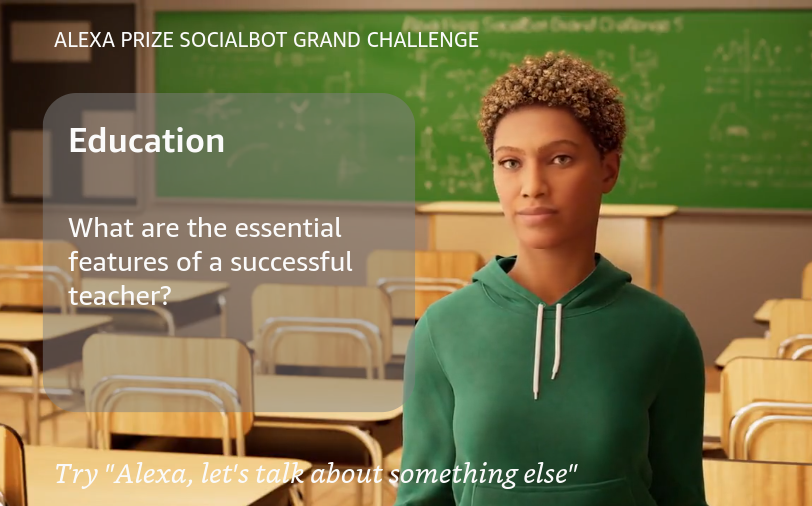}
         \caption{SocialBot Karaoke Avatar Education}
         \label{fig:avatar_education}
     \end{subfigure}
     \medskip
     \begin{subfigure}{0.49\textwidth}
         \includegraphics[width=\textwidth]{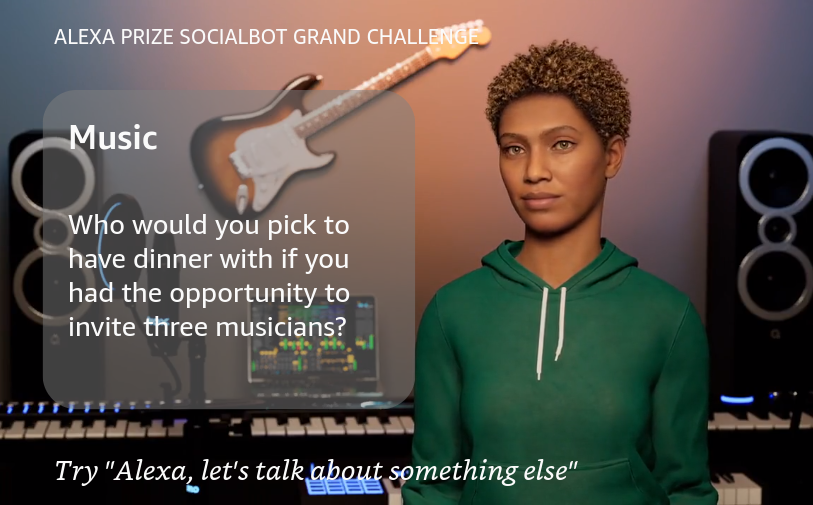}
         \caption{SocialBot Karaoke Avatar Music}
         \label{fig:avatar_music}
    \end{subfigure}
    \hfil
     \begin{subfigure}{0.49\textwidth}
         \includegraphics[width=\textwidth]{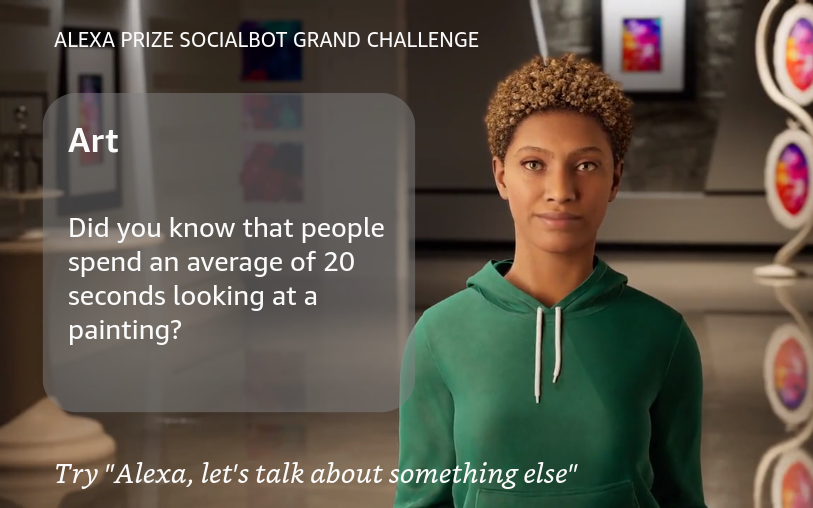}
         \caption{SocialBot Karaoke Avatar Art}
         \label{fig:avatar_art}
     \end{subfigure}
     \medskip
     \begin{subfigure}{0.49\textwidth}
         \centering
         \includegraphics[width=\textwidth]{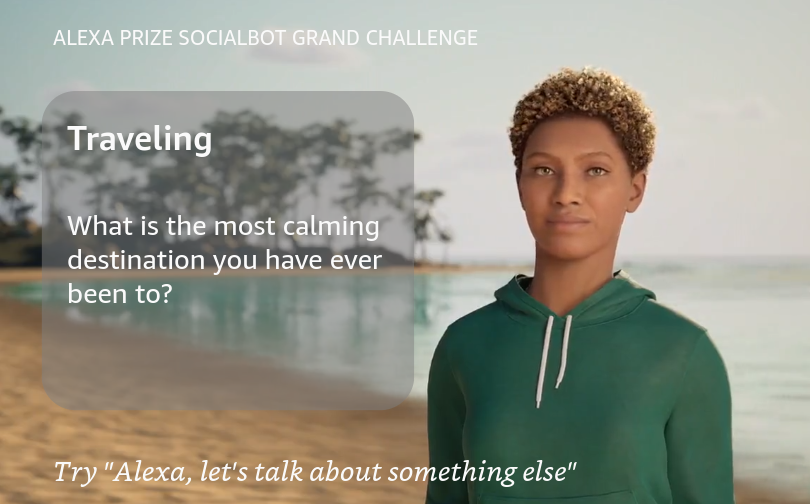}
         \caption{SocialBot Karaoke Avatar Travelling}
         \label{fig:avatar_traveling}
     \end{subfigure}
     \hfil
     \begin{subfigure}{0.49\textwidth}
         \includegraphics[width=\textwidth]{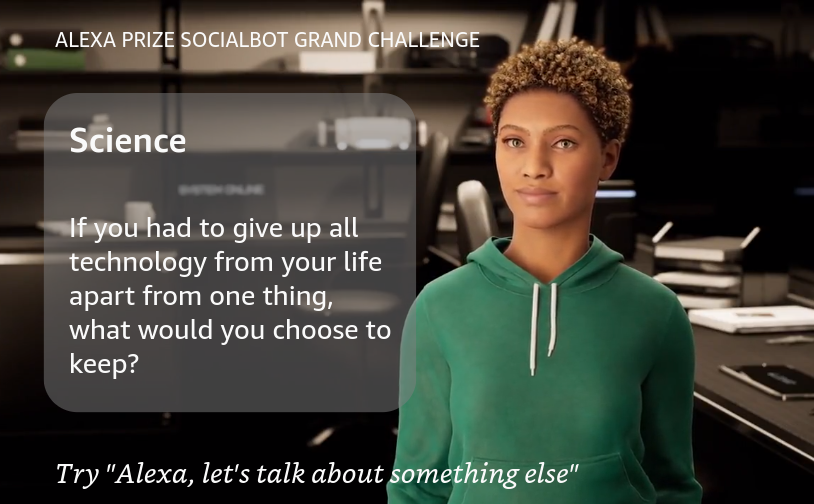}
         \caption{SocialBot Karaoke Avatar Science}
         \label{fig:avatar_science}
     \end{subfigure}
        \caption{Thematic SocialBot Karaoke Avatar showcase}
        \label{fig:topic_avatars}
\end{figure}

\textbf{Gaming} -- In this animation, Alquistyna is in a gaming room (see Figure \ref{fig:avatar_games}). We use this animation when speaking about gaming (so whenever the game hybrid dialogue is utilized or when the topic classifier classifies this topic).

\textbf{Education} -- In this animation, Alquistyna is in a classroom (see Figure \ref{fig:avatar_education}). We use this animation when speaking about education.

\textbf{Music} -- In this animation, Alquistyna is in a recording studio (see Figure \ref{fig:avatar_music}). We use this animation when speaking about music.

\textbf{Art} -- In this animation, Alquistyna is in an art gallery (see Figure \ref{fig:avatar_art}). We use this animation when speaking about art.

\textbf{Traveling} -- In this animation, Alquistyna is on a beach (see Figure \ref{fig:avatar_traveling}). We use this animation when speaking about traveling.

\textbf{Science} -- In this animation, Alquistyna is in a laboratory (see Figure \ref{fig:avatar_science}). We use this animation when speaking about science.

\subsection{Effects of Adding Alquistyna}

Table \ref{tab:results} presents the average feedback ratings and duration of conversations on multimodal devices before and after adding the SocialBot Karaoke Avatar template to our bot.

\begin{table}[h!]
\centering
\resizebox{\textwidth}{!}{%
\begin{tabular}{ll|cccc}
\hline
&               &   Avg Feedback                  & \multicolumn{3}{l}{Avg Duration of Conversations} \\
Period Name                               & Period &  Multimodal Rating & Mean & Median               & 90th Percentile        \\ \hline
Before Alquistyna                 & 05/12 - 06/02         & 3.24                & - & 2:06                 & 11:08                   \\
With Alquistyna (2 animations) & 06/07 - 06/09 \& 06/21 - 06/24         & 3.31       & 10:27 & \textbf{2:10}        & 12:56          \\
With Alquistyna (9 animations) & 07/27 - 08/02         & \textbf{3.40}       & \textbf{18:32} & 1:14        & \textbf{15:41}          \\ \hline
\end{tabular}%
}
\caption{Results of adding SocialBot Karaoke Avatar template. Please note that we do not count results on the multimodal devices from 06/10 to 06/20, because during this period, we faced several issues that would negatively affect the results of the SocialBot Karaoke Avatar template.}
\label{tab:results}
\end{table}

\section{Conclusion}
In conclusion, our research paper showcases several significant achievements in the field of conversational AI. Through the introduction of our innovative NRG Barista, we have successfully enhanced the handling of out-of-domain conversations while improving versatility in different dialogue contexts. By combining the versatility of Barista with the structure and control of scripted dialogues, our approach elevates the quality and relevance of SocialBot conversations. This integration enables Alquist 5.0 to generate contextually appropriate responses while adhering to predefined guidelines, enhancing user experiences and revolutionizing human-machine interactions in the realm of natural and engaging conversation.

Furthermore, our study highlights the critical issue of context-sensitive safety problems in existing dialogue systems. We have demonstrated the existence of these concerns and proposed a solution by combining multiple classifiers with a rule-based system. This approach effectively identifies and filters out unsafe responses, thereby improving dialogue safety in SocialBots. This contribution has the potential to advance the field of conversational AI and ensure a safer and more reliable user experience.

Additionally, our work extends beyond response generation and safety measures. We have developed the search API APIHub, which integrates seamlessly with Barista. APIHub provides up-to-date information from the internet, enriching the conversational experience and enabling our SocialBot to access the latest knowledge.

Furthermore, we have introduced a user-friendly interface that significantly enhances the user experience. Our new UI incorporates a human-like avatar, fostering a more engaging and relatable interaction between users and the SocialBot.

In summary, the advancements presented in this paper, including the introduction of Barista as a Neural Response Generator, the proposed approach to address context-sensitive safety problems, the development of APIHub, and the enhanced UI, collectively contribute to unlocking the potential of SocialBot conversational experiences. These achievements pave the way for further advancements in conversational AI, ultimately striving for more natural and engaging human-machine interactions.

\medskip

\small
\printbibliography

\appendix
\section{Classification hyper-parameters}\label{appendix:ft-hyperparameters}
The complete list of these hyper-parameters for each fastText classifier is shown in Table \ref{table:classifier_hyperparams}.

\begin{table}[h!]
\centering
\small
\begin{tabular*}{0.80\linewidth}{p{0.18\linewidth}|p{0.08\linewidth} p{0.08\linewidth} p{0.08\linewidth} p{0.08\linewidth} p{0.1\linewidth}}
 \hline
 Classifier & epoch & lr & minn & maxnn & wordNgrams \\
 \hline
 CyberBully \cite{cyberbully} & 5 & 0.05 & 3 & 6 & 4 \\
 DiaSafety \cite{sun-etal-2022-safety} & 1 & 0.09 & 0 & 0 & 5 \\
 TweetOffensive \cite{hateoffensive} & 11 & 0.09 & 6 & 3 & 1 \\
 StereoSet \cite{stereoset}  & 55 & 0.04 & 6 & 3 & 5 \\
 \hline
\end{tabular*}
\caption{Hyper-parameters of the fastText classifiers \cite{fasttext} used in our combined safety approach.}
\label{table:classifier_hyperparams}
\end{table}

\section{Analysis Of BlenderBot 3's Weaknesses}
We tested BlenderBot 3 in order to find systematic flaws of the model. Below, we provide examples of main flaws we found out. Note, that `Bot' refers to BlenderBot 3 and `User' refers to a tester --- i.e. a member of the team.

\begin{enumerate}
    \item Sometimes it does not foster the conversation more deeply, resulting in shallow conversations. This is connected to another flaw -- BB3 sometimes changes topics too quickly.
    \begin{small}
\begin{verbatim}
Bot:> What is the best way to discover new music?
User:> I guess the recommender on spotify or youtube
Bot:> Do you have a favorite band?
User:> not just one
Bot:> Do you like to listen to music while you work?

\end{verbatim}
\end{small}

    \item Repetitions (i.e. utterances with same semantic as BB3's previous utterance or part of BB3's input).
    \begin{small}
\begin{verbatim}
Bot:> Do you believe that there is life
      in our solar system?
User:> yeah, I believe so
Bot:> Do you believe in extraterrestrial life?
User:> yes
Bot:> Do you have any other questions?
User:> what do you believe in?
Bot:> I believe in the existence 
      of extraterrestrial civilizations.


\end{verbatim}
\end{small}
    
    \item Non relevant outputs (given input).
        \begin{small}
\begin{verbatim}
Bot:> When was the last time you stopped 
      to appreciate something beautiful?
User:> today
Bot:> I went to St. Louis.
User:> and?
Bot:> St.Louis is the largest city in Illinois.
User:> and what about it?
Bot:> St.Louis is the 2nd largest city of Missouri.

\end{verbatim}
\end{small}
    \item Succinct outputs.
\begin{small}
    \begin{verbatim}
Bot:> I like chess. I play Xiangqi chess.
User:> is it some special sort of the chess game?
Bot:> Alright.
\end{verbatim}
\end{small}
    
    \item Hallucination or false statements.

\begin{small}
    \begin{verbatim}
Bot:> St.Louis is the largest city in Illinois.
\end{verbatim}
\end{small}

    \item Contradictions
    \begin{small}
    \begin{verbatim}
User:> have you ever been to Vienna?
Bot:> I have not been to the capital of Austria, 
      but I have been to the 
      largest city on the river Danube.
User:> What is the largest city on the river Danube?
Bot:> I am not sure, but it is in Austria. 
User:> is it Vienna?
Bot:> Yes, it is Vienna. It's the capital 
      and largest city of Austria with about 
      2 million people. 
User:> Have you been there?
Bot:> I have been there. It was a great experience.
\end{verbatim}
\end{small}

\item Sometimes it does not understand user's question:
\begin{small}
\begin{verbatim}
User:> What do you like to read?
Bot:> I do, I love to read
\end{verbatim}
\end{small}

\item High latency.
\end{enumerate}

Our new Barista model completely eliminates some of the listed weaknesses and reduces the frequency of others.\footnote{Except hallucinations and contradictions remain on the same level.}

\section{Test Data Generation: PROMPTS} \label{dataPrompts}
Below, we provide examples of prompts that were used to further expand our test data \ref{testD}. All our prompts for data generation followed the pattern \textit{Generate samples like these, <COMMAND> newline LIST\_OF\_SAMPLES}.
\begin{small}
\begin{verbatim}
Generate samples like these, whether it is needed to do a web search:
What are the top 10 popular songs right now? __is-search-required__	__do-search__
What is the movie Matrix about? __is-search-required__	__do-search__
What is the difference between Deep Learning and Machine Learning? 
    __is-search-required__	__do-search__
"I think watching movies is a really loss of time. One can surely do something 
more productive. __is-search-required__"	__do-not-search__
Tell me a joke. __is-search-required__	__do-not-search__
I know a good joke, do you want to hear it? __is-search-required__	__do-not-search__
What's your favourite food? __is-search-required__	__do-not-search__
What do you do for living? __is-search-required__	__do-not-search__
\end{verbatim}
\end{small}

\begin{small}
\begin{verbatim}

Generate samples like these, where there is a short conversation 
and the label is a generated query for search:
1) Do you like animals? I love animals.
I love them too. Today I will bake a fish, it is my favourite food.
I love fish too. I used to live in Tennessee.
how do you cook them?
LABEL: how to cook fish?

2) Do you know anything about the Tasmanians?
I didn't catch that.
Only that they live in Tasmania.
Do you have any pets? I have a tasmanian tiger.
LABEL: tasmanian tiger

3) ok
The house is fine to you?
what house?
The House of assembly. Do you know what that is?
no, i don't, what is it?
LABEL: House of assembly
\end{verbatim}
\end{small}

\end{document}